%% file: arxiv.tex
\documentclass[10pt,twocolumn,letterpaper]{article}

\usepackage[pagenumbers]{cvpr} %

\input{preamble}
\usepackage{multirow}
\usepackage{multicol}
\usepackage{booktabs}
\usepackage{xcolor}         %
\usepackage{wrapfig} %
\usepackage{caption}
\usepackage{subcaption}
\usepackage{graphicx}
\usepackage{colortbl}
\usepackage[rightcaption]{sidecap}
\usepackage{float}

\usepackage{xr}
\usepackage[accsupp]{axessibility} 
\definecolor{cvprblue}{rgb}{0.21,0.49,0.74}
\usepackage[pagebackref,breaklinks,colorlinks,allcolors=cvprblue]{hyperref}

\definecolor{skip}{HTML}{F2ACCA}
\definecolor{reuse}{HTML}{9CCEA7}
\definecolor{adapt}{HTML}{FEB24C}
\definecolor{new}{HTML}{9EC9E2}

\definecolor{tabred}{HTML}{d62728}
\definecolor{tabblue}{HTML}{1f77b4}

\definecolor{vitb}{HTML}{DAE8FC}
\definecolor{deit}{HTML}{D5E8D4}
\definecolor{highlight}{HTML}{F4E3FD}

\newcommand\newsubcap[1]{\phantomcaption%
       \caption*{\figurename~\thefigure\thesubfigure: #1}}

\def\paperID{18222} %
\def\confName{CVPR}
\def\confYear{2026}

\title{CHEEM: Continual Learning by Reuse, New, Adapt and Skip \--- \\ A Hierarchical Exploration-Exploitation Approach}

\author{Chinmay Savadikar \\
North Carolina State University\\
\texttt{csavadi@ncsu.edu}\\
\and
Michelle Dai \\
Johns Hopkins University\\
\texttt{mdai12@jh.edu} \\
\and
Tianfu Wu \\
North Carolina State University\\
\texttt{tianfu\_wu@ncsu.edu}
}

\begin{document}
\maketitle

\setlength{\abovecaptionskip}{1pt}
\setlength{\belowcaptionskip}{-6pt}
\setlength{\belowdisplayskip}{1pt} \setlength{\belowdisplayshortskip}{1pt}
\setlength{\abovedisplayskip}{1pt} \setlength{\abovedisplayshortskip}{1pt} 

\input{sec/abstract_arxiv}    
\input{sec/intro_arxiv}

{
    \small
    \bibliographystyle{ieeenat_fullname}
    \bibliography{main}
}

\onecolumn
\maketitle

\appendix
\section*{Appendix}
\input{sec/appendix_arxiv}

\end{document}

%% file: sec/abstract_arxiv.tex
\begin{abstract}
To effectively manage the complexities of real-world dynamic environments, continual learning must incrementally acquire, update, and accumulate knowledge from a stream of tasks of different nature—without suffering from catastrophic forgetting of prior knowledge. While this capability is innate to human cognition, it remains a significant challenge for modern deep learning systems.
At the heart of this challenge lies \textit{the stability-plasticity dilemma}: the need to balance leveraging prior knowledge, integrating novel information, and allocating model capacity adaptively based on task complexity and synergy. 
In this paper, we propose a novel exemplar-free class-incremental continual learning (ExfCCL) framework that addresses these issues through a \textbf{Hierarchical Exploration-Exploitation (HEE) }approach.
The core of our method is a HEE-guided efficient neural architecture search (HEE-NAS) that enables a learning-to-adapt backbone via four primitive operations—reuse, new, adapt, and skip—thereby serving as an internal memory that dynamically updates selected components across streaming tasks. 
To address the task ID inference problem in  ExfCCL, we exploit an external memory of task centroids proposed in the prior art. We term our method \textbf{CHEEM} (Continual Hierarchical-Exploration-Exploitation Memory). CHEEM is evaluated on the challenging MTIL and VDD benchmarks using both Tiny and Base Vision Transformers and a proposed holistic \textbf{Figure-of-Merit (FoM) metric}. It significantly outperforms state-of-the-art prompting-based continual learning methods, closely approaching full fine-tuning upper bounds. Furthermore, it learns adaptive model structures tailored to individual tasks in a semantically meaningful way. Our code is available at \url{https://github.com/savadikarc/cheem}.
\end{abstract}

%% file: sec/intro_arxiv.tex
\vspace{-6mm}
\section{Introduction}
\label{sec:intro}
Developing continual learning machines is a key objective in Artificial Intelligence (AI), aiming to replicate human-like adaptability and the ability to learn-to-learn, enabling proficiency in streaming tasks. Despite their advances, state-of-the-art Deep Neural Networks (DNNs) still lack true biological intelligence in the realm of continual learning from streaming tasks in dynamic environments, which requires the continual acquisition, update, and accumulation of knowledge while mitigating catastrophic forgetting of previous tasks~\citep{mccloskey,thrun}, referring to the stability-plasticity trade-off.

\begin{figure*}
    \centering
    \begin{minipage}[b]{0.75\linewidth}
        \includegraphics[width=1.0\linewidth]{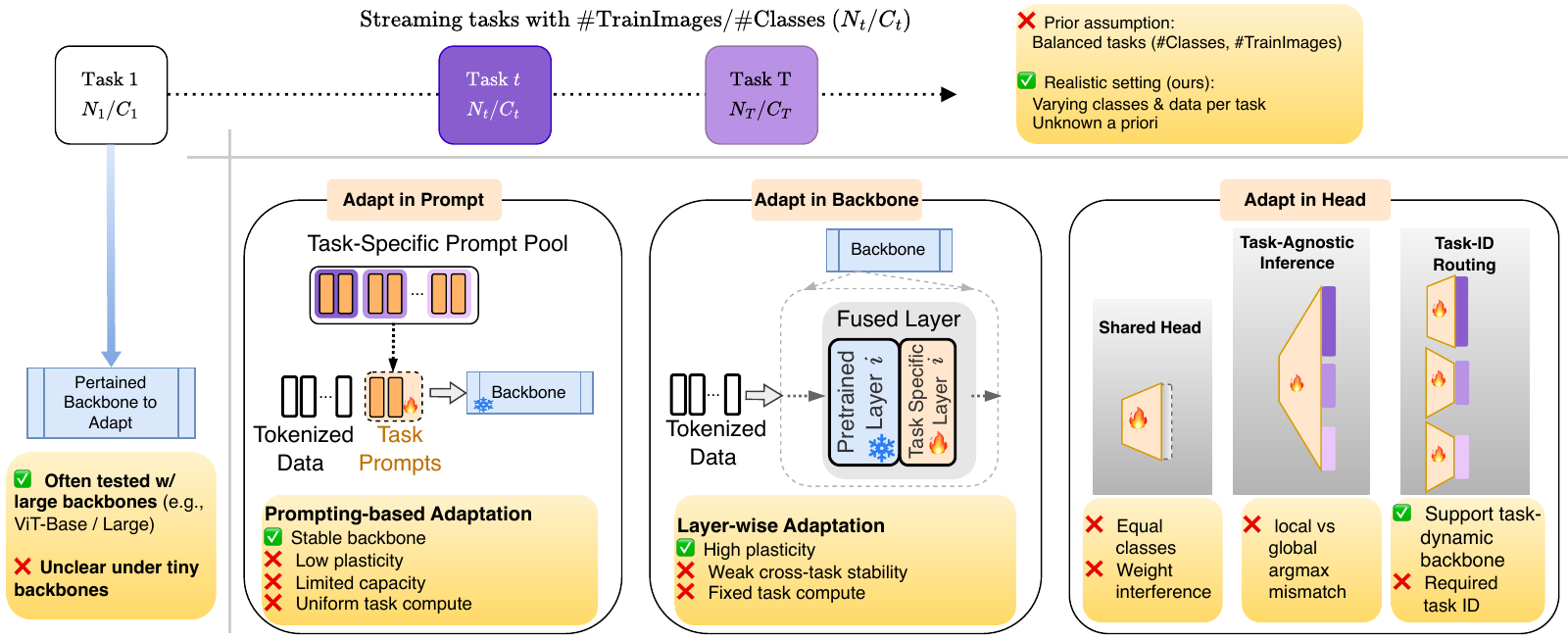}
    \end{minipage}\hspace*{2mm}%
    \begin{minipage}[b]{0.24\linewidth}
        \captionof{figure}{
        \textbf{Taxonomy of ExfCCL design choices.} Existing methods typically adapt either the \emph{prompt}~\citep{learning-to-prompt,dualprompt,s-prompts,coda-prompt} or the \emph{backbone} (often layer-wise~\citep{supsup,meta-attention,piggyback,ll,clr,pool-of-adapters,nettailor}), together with different strategies in the \emph{head} adaptation, each with trade-offs in stability, plasticity, inference assumptions, and compute adaptivity.
        See text for detail.
        }
        \label{fig:prior}
    \end{minipage}\vspace{-2mm}
\end{figure*}

\begin{figure*} [t]    
    \begin{subfigure}{\textwidth}
        \centering
        \includegraphics[width=.9\linewidth]{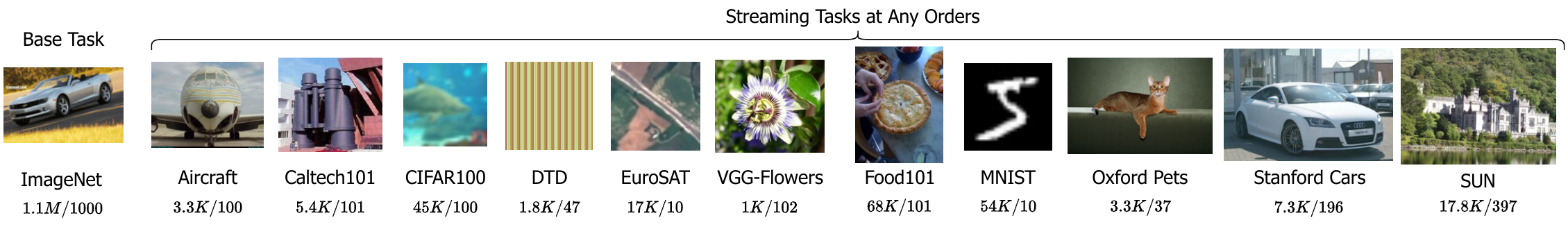}
        \caption{\textbf{The MTIL benchmark}~\citep{mtil} consisting of tasks of different nature with {\tt \#training images/\#classes} significantly varying across different tasks. }
        \label{fig:teaser-mtil}
    \end{subfigure}
    \vspace{0.1mm}
    
    \begin{subfigure}{\textwidth}
        \centering
        \includegraphics[width=.85\linewidth]{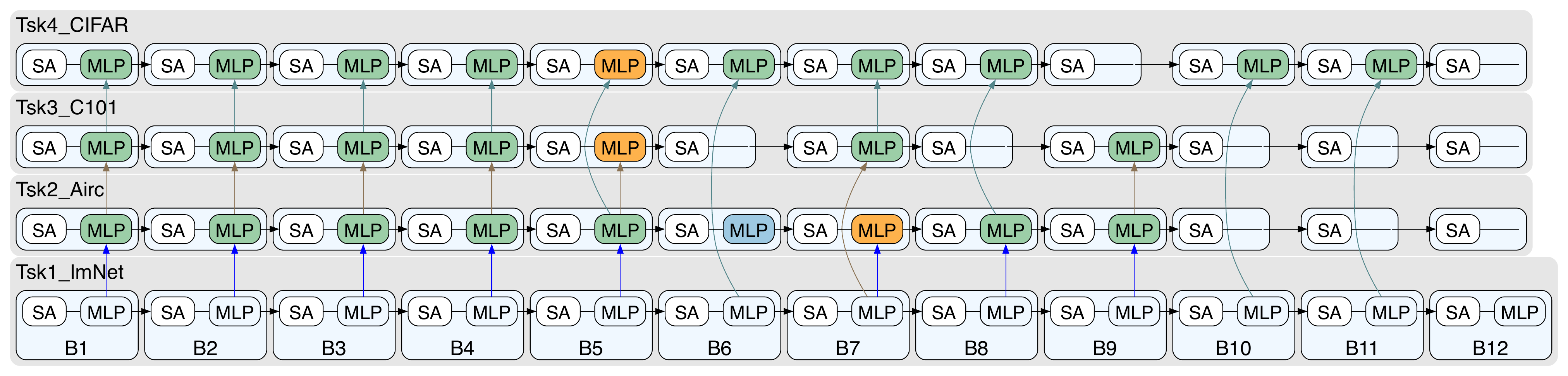}
        \caption{From \textbf{ViT-Base trained on Tsk1\_ImNet} (with blocks B1 to B12), our CHEEM learns sensible task-tailored models that reflect the task complexity. For example, when learning Caltech 101 (Tsk3\_C101), CHEEM learns to {\tt Skip} 5 MLP blocks and \colorbox{reuse}{\tt Reuse} most of the architecture. On the contrary, when learning FGVC Aircraft (Tsk1\_Airc), which is a more complex task with larger shift from ImageNet due to its fine-grained nature, CHEEM learns to \colorbox{adapt}{\tt Adapt} the ImageNet parameters in Block 7, adds a \colorbox{new}{\tt New} operation in Block 6, and {\tt Skips} the last 3 MLP blocks. See text for details. Full structure reproduced in Fig. \ref{fig:vit-b-ee-full} in the supplementary.}
        \label{fig:teaser-vit-b}
    \end{subfigure}
    \vspace{0.1mm}
    
    \begin{subfigure}{\textwidth}
        \centering
        \includegraphics[width=0.85\linewidth]{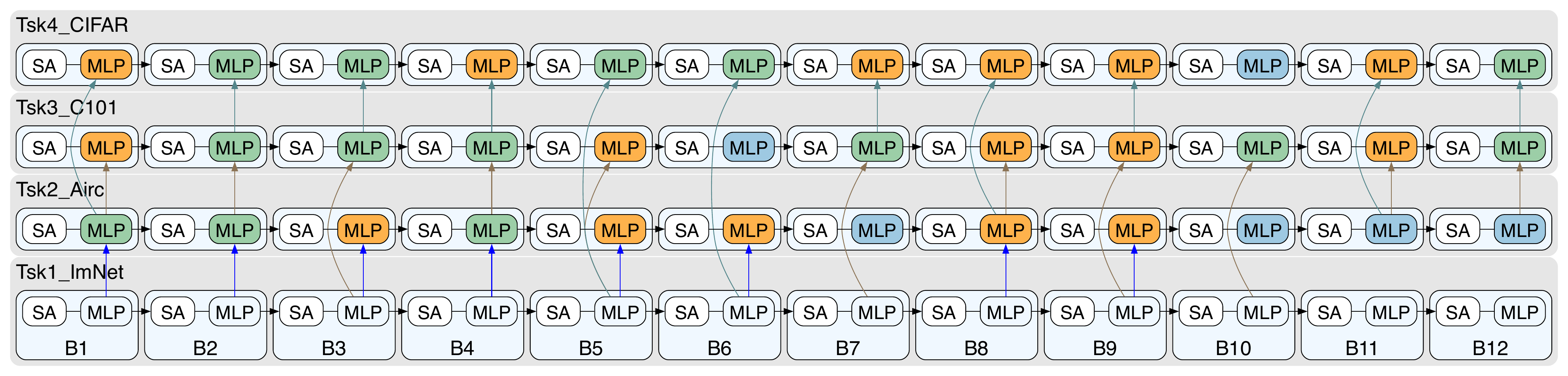}
        \caption{From \textbf{DEiT-Tiny trained on Tsk1\_ImNet} (with blocks B1 to B12), our CHEEM learns to use multiple \colorbox{adapt}{\tt Adapt} and \colorbox{new}{\tt New} operations, without {\tt Skip} operations selected, sensibly different from those with more {\tt Skip} and less \colorbox{new}{\tt New} operations learned based on the stronger ViT-Base model. Full structure reproduced in Fig. \ref{fig:diet-tiny-ee} in the supplementary.}
        \label{fig:teaser-deit-t}
    \end{subfigure}
    \vspace{0.3mm}
    
    \caption{Examples of CHEEM learning task-tailored models.}
    \label{fig:teaser}\vspace{-2mm}
\end{figure*}

Recently, continual learning using Vision Transformers (ViTs)~\citep{vit} has witnessed promising progress - particularly in the Exemplar-free Class Incremental Continual Learning (ExfCCL) setting, where the raw data (or latent features of samples) of old tasks are not available in learning a new task, and task IDs of testing samples are unknown at inference. 
Fig.~\ref{fig:prior} shows an overview of the prior art in ExfCCL. There are four key aspects of interest in this paper: 

\textbf{i) What benchmarks to use?} Many existing works focus on unrealistic balanced benchmarks such as Split ImageNet-R benchmark \cite{learning-to-prompt} with equal number of classes per streaming tasks, as well as equal number of training images. In the literature, challenging benchmarks such as VDD~\cite{vdd} that has significantly varying number of classes and training images per task has been used in early works like ~\cite{learn-to-grow} in the  task-incremental setting. More recently, benchmarks of similar nature to VDD such as MTIL~\cite{mtil} have been proposed to test ExfCCL in more practical real-world scenarios (see examples in Fig.~\ref{fig:teaser-mtil}). We adopt these benchmarks and focus on both VDD and MTIL in this paper. 

\textbf{ii) How to adapt ViT backbone to address the stability and plasticity challenge?} Task $1$ (e.g., ImageNet-1k) is typically assumed to train a ViT sufficiently well~\citep{learning-to-prompt,dualprompt,s-prompts,coda-prompt}. There are two main backbone adaptation strategies:  \textit{prompting-based methods }that retain strong stability but lack plasticity by keeping the ViT backbone frozen throughout all streaming tasks, and resorting to learning task-specific prompts or prefixes~\citep{learning-to-prompt,dualprompt,s-prompts,coda-prompt}, and \textit{parameter-tuning methods} that maintain plasticity but lack task-synergy stability by adapting all layers in a ViT using either parameter-masking~\citep{supsup,meta-attention,piggyback}, output addition~\citep{ll,clr,pool-of-adapters,nettailor}, or adapter tuning \cite{sd-lora,inf-lora}. 
More importantly, both strategies often overlook and fail to address a critical aspect: \textbf{how to adapt computation to be task-difficulty/synergy-aware?} 
Prompting-based methods emphasize {\tt Reuse} of every layer in a ViT, while parameter-tuning methods utilize {\tt Adapt} of every layer. Both keep ViT backbone architecture frozen, and thus they either retain the same computation across tasks or increase it when new task-specific prompts appended in the input, regardless of task difficulty. 
There are two critical components underexplored in the prior art: One is a {\tt Skip} operation that bypasses certain layers in ViT when a task is a relatively easy one, enabling smaller backbones for easy tasks. The other is a {\tt New} operation that introduces a totally new layer to substitute an existing one when a task is significantly different from previous ones, enabling novel information integration quicker and better than {\tt Adapt} (that is constrained by existing pretrained weights). In this paper, we propose a method of learning the four operations ({\tt Reuse, Adapt, New} \& {\tt Skip}) to best balance the stability and plasticity in addressing challenges of ExfCCL in the VDD and MTIL benchmarks, which leads to task-difficulty/synergy-aware dynamic backbones.  

\begin{figure}%
    \centering
    \includegraphics[width=0.8\linewidth]{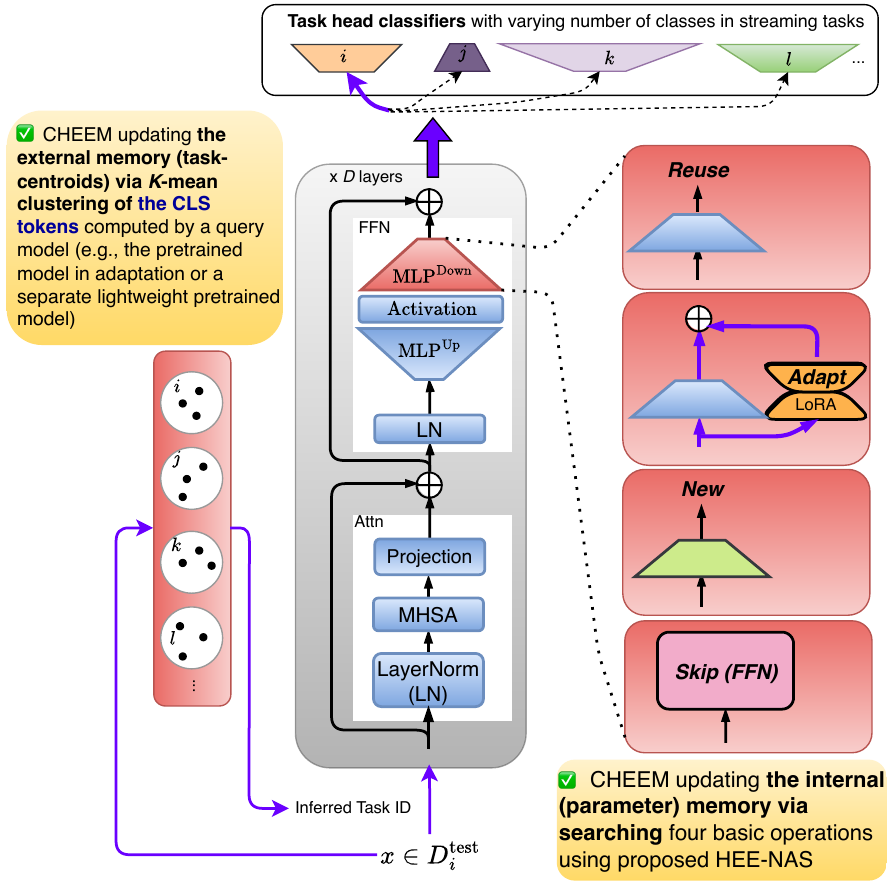}
    \caption{\textbf{Illustration of the proposed CHEEM.} 
     A pretrained and frozen ViT model such as ViT-Base~\citep{vit} or DEiT-Tiny~\citep{deit} is  structurally and dynamically updated to learn internal (parameter) memory for streaming tasks in continual learning, and is also used in maintaining the external task-centroid memory of CHEEM.   
    CHEEM learns the internal parameter memory 
    for a selected component such as the MLP$^{\text{Down}}$ layer. We also test placing CHEEM at the projection layer in the `Attn' block. 
    }
    \label{fig:flow} \vspace{-4mm}
\end{figure}

\textbf{iii) How to learn task head classifiers?} There are three different designs with corresponding underlying assumptions. \textit{A shared task head } is used when all tasks are assumed to have the exact same number of classes in those unreal balanced benchmarks, which also entails strong weight consolidation in training to overcome catastrophic forgetting in ExfCCL. 
\textit{A task-agnostic head } can eliminate the need of unreal assumption of equal class numbers across tasks and is designed to handle no explicit task-ID inference. However, it suffers from the discrepancy between training and inference. During training, task ID is known, only local-argmax is used in computing the loss, while in inference, without task ID of a testing sample, the global-argmax is used to predict the class. The consensus between local-argmax and global-argmax is very difficult to preserve in ExfCCL, for which we show a theoretical justification in the supplementary (Sec.~\ref{sec:growing-head-analyses}). Our experimental results also empirically reflect the difficulty of this task-agnostic head design. \textit{Task-specific heads} do not suffer from those issues, but entail explicit task ID inference of testing samples~\cite{s-prompts}. In this paper, we adopt task-specific head design, which is consistent with our proposed dynamic backbones, both entailing task ID inference.

\textbf{iv) What capacity of ViT backbone to start with?} Most of existing work start with high-capacity ViT backbones such as ViT-Base/Large, which leaves ExfCCL with tiny backbones underexplored. ExfCCL with tiny backbones is useful in practice for the deployment of continual learning on edge devices. ExfCCL with tiny backbones also challenges the prompting-based methods since they are upper-bounded by the capacity of the backbone, and advocates the need of introducing {\tt New} operations. We test both base and tiny models with different yet meaningful task-tailored backbones learned (Fig.~\ref{fig:teaser-vit-b} and~\ref{fig:teaser-deit-t}).

 Fig.~\ref{fig:flow} illustrates our proposed method. We view ExfCCL from continual memory learning perspective, consisting of our proposed novel internal parameter memory learning and external task centroid memory learning adopted from the prior art~\cite{s-prompts}. We term our proposed method \textbf{CHEEM} (\textit{Continual Hierarchical-Exploration-Exploitation Memory}) in which \textbf{a new task learns to automatically  reuse/adapt modules from previous similar tasks, to introduce new modules when needed, or to skip some modules when it appears to be an easier task} (see Figs.~\ref{fig:teaser-vit-b} and~\ref{fig:teaser-deit-t}). We propose a hierarchical exploration-exploitation (HEE) sampling based neural architecture search (NAS) method for learning the internal memory.

    To ensure NAS is computationally efficient, and retain the stability of the backbone to account for tasks in streams that have little training data, we select two components in a ViT block: the down projection layer (MLP$^{\text{Down}}$) in the FFN and  the projection layer after the MHSA, to be plastic to maintain \textbf{the internal parameter memory in a compute-budget controllable way} using: 
    \begin{itemize}[leftmargin=*,noitemsep,topsep=0pt]
        \item \colorbox{reuse}{\tt{Reuse}}: Facilitates similar tasks sharing layers for knowledge transfer in continual learning. 
        \item \colorbox{new}{\tt {New}}: Explores new features for handling tasks that are dissimilar to previous tasks. It enables learning-to-grow the backbone to be skilled at streaming tasks. 
        \item \colorbox{adapt}{\tt {Adapt}}: Utilizes LoRA~\citep{lora}, inducing task synergies in ExfCCL in a parameter-efficient way.        
        \item \colorbox{skip}{\tt {Skip}}: Skips the entire FFN block (when the MLP$^\text{Down}$ is used) or the entire MHSA block accordingly. It can thus induce much simpler backbones for relatively easier tasks in a learning-to-prune manner, especially when a strong backbone such as ViT-Base is used (e.g., from ImageNet to MNIST). 
    \end{itemize}

In experiments, to account for the five metrics (average accuracy, average forgetting, average parameter increase and average compute) holistically in ExfCCL, we propose a holistic \textbf{Figure of Merit (FoM)} based metric to compare CHEEM with baseline methods.  Our CHEEM is tested on two challenging benchmarks (MTIL~\citep{mtil} and VDD~\citep{vdd}) using both ViT-Base~\citep{vit} and DEiT-Tiny~\citep{deit}  and obtains significantly better performance than prompting-based methods~\citep{coda-prompt,dualprompt,learning-to-prompt,s-prompts,diki}. Our CHEEM's performance is close to the upper-bound performance using either task-to-task full fine-tuning or task-to-task LoRA based fine-tuning, demonstrating its effectiveness. The learned task-tailored backbones are also sensible, and result in much less overall computing cost across all tasks compared to prompting based methods.

\section{Related Work and Our Contributions}

For exemplar-free continual learning, \textit{Regularization Based approaches} explicitly control the plasticity of the model by preventing the parameters of the model from deviating too far from their stable values learned on the previous tasks when learning a new task~\citep{ewc,what-not-to-forget,selfless-sequential-learning,podnet,variational-continual-learning,kirkpatrick-overcoming,lwf,synaptic-intelligence,progress-and-compress}. These approaches aim to balance the stability and plasticity of a fixed-capacity model.
\textit{Dynamic Models} aim to use different parameters for each task to eliminate the use of stored exemplars. Dynamically Expandable Network~\citep{dynamic-expandable-nets} adds neurons to a network based on learned sparsity constraints and heuristic loss thresholds. PathNet~\citep{pathnet} finds task-specific submodules from a dense network, and only trains submodules not used by other tasks. Progressive Neural Networks~\citep{pnn} learn a new network per task and adds lateral connections to the previous tasks' networks. ~\citep{vdd} learns residual adapters which are added between the convolutional and batch normalization layers. \citep{network-of-experts} learns an expert network per task by transferring the expert network from the most related previous task. The L2G~\citep{learn-to-grow} uses Differentiable Architecture Search (DARTS)~\citep{darts} to determine if a layer can be reused, adapted, or renewed for a task, which is tested for ConvNets and the learning-to-grow operations are applied uniformly at each layer in a ConvNet. Our method is motivated by the L2G method, but with substantially significant differences.  

Recently, there has been increasing interest in continual learning using Vision Transformers~\citep{learning-to-prompt,dualprompt,meta-attention,pool-of-adapters,dytox,towards-exemplar-free-continual-learning-vits,improving-vits,continual-obj-det-kd,memory-transformer,s-prompts,lifelong-vision-transformer,d3former,Gao_2023_ICCV}. \textit{Prompt Based approaches} learn external parameters appended to the data tokens that encode task-specific information useful for classification~\citep{learning-to-prompt,s-prompts,dytox,coda-prompt,dualprompt,diki}. Our proposed method is complementary to prompting-based methods. \textit{Adapter-based approaches} learn task-specific or universal adapters \cite{tuna}, or use a mixture-of-experts style adapters \cite{moeadapters4cl}.

\textbf{Our Contributions}. 
This paper makes  three main contributions to the field of ExfCCL using ViT:  
(i) It poses ExfCCL as a problem of learning two decoupled continual memory in ViT, the external task-centroid memory and the internal parameter memory.
(ii) It presents a hierarchical task-synergy exploration-exploitation sampling based NAS method for maintaining the internal memory by learning task-aware dynamic models continually with respect to four operations: {\tt Reuse}, {\tt Adapt}, {\tt New} and {\tt Skip}, to mitigate  catastrophic forgetting. 
(iii) It shows state-of-the-art performance on two challenging benchmarks (MTIL and VDD) in terms of a proposed Figure of Merit (FoM) metric, with sensible task-tailored model structures automatically learned.

\section{Our Proposed CHEEM}\label{sec:approach}
This section presents details of our proposed CHEEM. 
We start with a vanilla $D$-layer ViT model (e.g., the 12-layer ViT-Base)~\citep{vit}.
As illustrated in Fig.~\ref{fig:flow}, we select two components in a Transformer block,  the MLP$^{\text{Down}}$ and the project layer after the MHSA to place the internal parameter memory (see Sec.~\ref{sec:place_cheem} in the supplementary).

\subsection{The Mixture-of-Experts Representation of Task-Synergy Internal Memory}\label{sec:learning_cheem}
\begin{figure}
    \centering
    \includegraphics[width=1.0\linewidth]{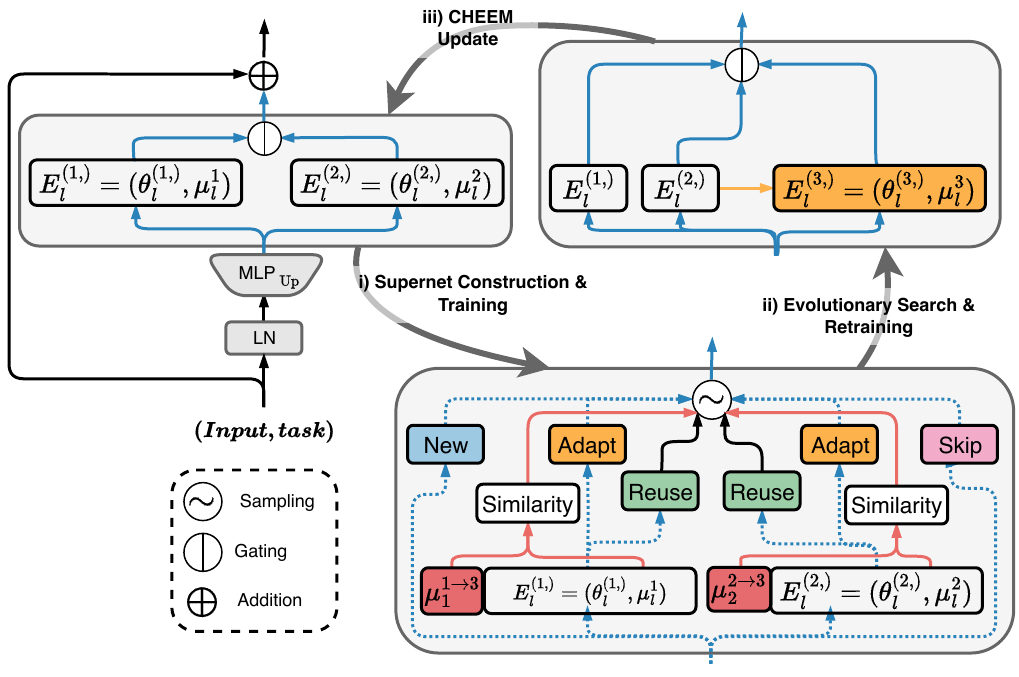}
    \caption{Illustration of CHEEM learning via NAS. }
    \label{fig:nas} \vspace{-3mm}
\end{figure}

The proposed internal memory of our CHEEM  is represented by a Mixture of Experts (MoEs). Starting with the ViT base model $F_1$, the internal memory at the $l$-th layer in  ViT  consists of a single expert defined by a tuple, 
\begin{equation}
{\tt E}_l^{(1,)}=(\theta_l^{(1,)}, \mu_l^{1}),    
\end{equation}
where the subscript represents the layer index and the list-based superscript shows which task(s) use this expert. $\theta_l^{(1,)}$ are the parameters of the projection layer or the MLP$^{\text{Down}}$ layer and $\mu_l^{1}\in R^d$ is the associated mean class-token ({\tt CLS}) pooled from the training dataset after the model is trained, which is task specific (as indicated by the superscript).  For example, if an expert is reused by another task (say, 3) in continual learning, we will have ${\tt E}_l^{(1,3,)}=(\theta_l^{(1,3,)}, \mu_l^{1}, \mu_l^{3})$.

As shown in Fig.~\ref{fig:nas},  for a new task $t$, learning to update CHEEM consists of three components: i) the Supernet construction (the parameter space of updating CHEEM), ii) the Supernet training (the parameter estimation of updating CHEEM), and iii) the target network selection and finetuning (the consolidation of the CHEEM for the task $t$).

\subsection{Supernet Construction}\label{sec:supernet_construction}

For clarity, we consider how the space of MoEs of the internal memory is constructed at a single layer $l$ for a new task with CHEEM placed at the MLP$^{\text{Down}}$ (projection) layer, assuming the current memory consists of two experts, $\{{\tt E}_l^{(1,)}, {\tt E}_l^{(2,)}\}$ (Fig.~\ref{fig:nas}, left). {The Supernet is constructed via:} 
\begin{itemize} [leftmargin=*,noitemsep,topsep=0pt]
    \item   \colorbox{reuse}{\tt{Reuse}}: Uses the MLP$^{\text{Down}}$ (projection) layer from an old task for the new task unchanged, exploiting task synergies during learning.
    \item   \colorbox{adapt}{\tt Adapt}: Introduces a new lightweight LoRA~\citep{lora} component, e.g., $\theta_l^{(3,)}=\theta_l^{(2,)}+B_l\cdot A_l$, where $B_l$ and $A_l$ are low-rank parameter matrices.
    \item  \colorbox{new}{\tt New}: Adds a new MLP$^{\text{Down}}$ (projection) layer, which enables the model to handle corner cases and novel situations.
    \item  \colorbox{skip}{\tt Skip}: Skips the entire FFN (MHSA) block, which encourages dynamically adjusting the model complexity based on the task complexity. 
\end{itemize}

The bottom of Fig.~\ref{fig:nas} shows the search space. The Supernet is constructed by {\tt reusing} and {\tt adapting} each existing expert at layer $l$, and adding a {\tt new} and a {\tt skip} expert. The newly added {\tt adapt} ($B_l, A_l$) by LoRA and projection layers will be trained from scratch using the data of a new task only. The right-top of Fig.~\ref{fig:nas} shows the {\tt Adapt} operation on top of ${\tt E}_l^{(2,)}$ is learned and added, ${\tt E}_l^{(3,)}=(\theta_l^{(3,)}, \mu_l^3)$ where $\mu_l^3$ is the mean {\tt CLS} token pooled for the task 3. %

\subsection{Supernet Training via HEE-NAS}\label{sec:supernet_train}

\begin{figure} %
    \centering
    \includegraphics[width=1.0\linewidth]{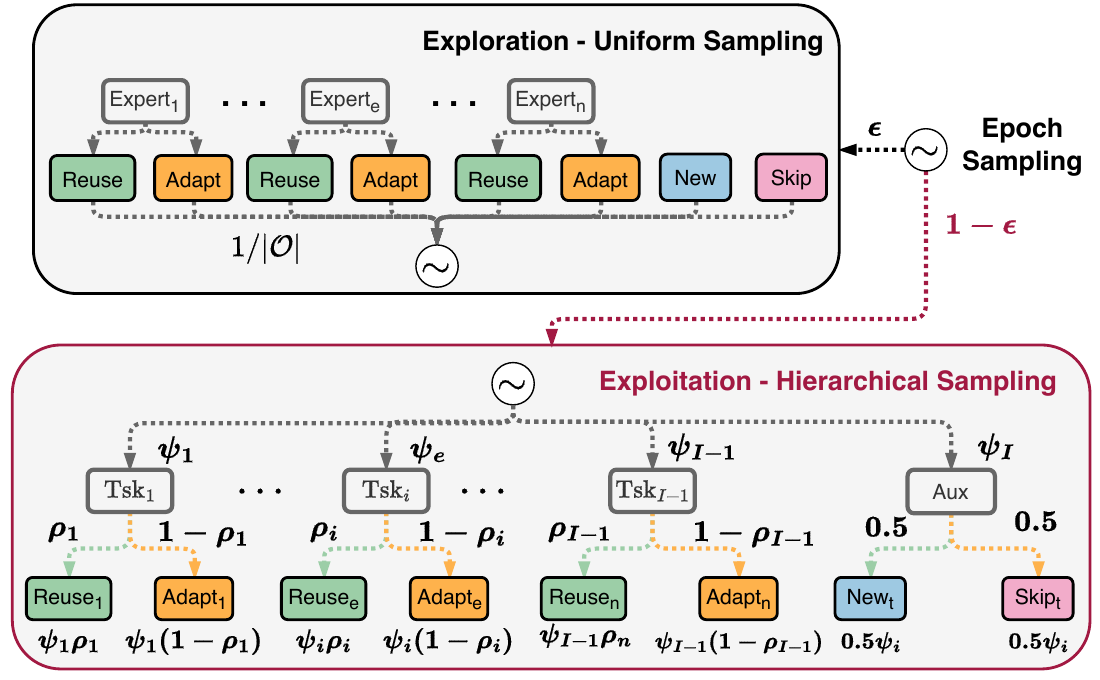}
    \caption{
    Illustration of the proposed HEE sampling based NAS.  It integrates the vanilla  exploration strategy (top) and the proposed exploitation strategy (bottom) with an epoch-wise scheduling.}
    \label{fig:nas-sampling}\vspace{-2mm}
\end{figure}

To train the Supernet constructed for a new task $t$, we build on the efficient SPOS method \citep{spos}. The vanilla SPOS trains a single-path sub-network from the Supernet by sampling an expert at every layer in each mini-batch. One key aspect is the sampling strategy. The vanilla SPOS method uses uniform sampling (i.e., the \textit{pure exploration} (PE) strategy, Fig.~\ref{fig:nas-sampling} top).
We propose an exploitation strategy (Fig.~\ref{fig:nas-sampling} bottom), which utilizes a hierarchical sampling method that forms the categorical distribution over the operations in the search space \textbf{explicitly based on task synergies computed based on the pooled task-specific {\tt CLS} tokens}.

Consider a new task $t$ with the training dataset $D^{train}_t$, with the current supernet consisting of $t-1$ task-specific target networks, we first run inference of the $t-1$ target networks on $D^{train}_t$ to pool initial {\tt CLS} tokens for each expert, e.g., $\mu_l^{1\rightarrow 3}$ and $\mu_l^{2\rightarrow 3}$ in the bottom of Fig.~\ref{fig:nas}. 
Consider one expert ${\tt E}_l^{(i,j,)}$ at the $l$-th layer which is shared by two previous tasks $i$ and $j$ with their mean {\tt CLS} tokens $\mu_l^i$ and $\mu_l^j$ respectively, we have the pooled {\tt CLS} tokens for the current task $t$,  $\mu_l^{i\rightarrow t}$ and $\mu_l^{j\rightarrow t}$, computed accordingly. The task similarity is computed by, 
\begin{equation}
    S_l^{i,t} = \texttt{NormCosine}(\mu_l^i, \mu_l^{i\rightarrow t}),
\end{equation}
where {\tt NormCosine}$(\cdot, \cdot)$ is the Normalized Cosine Similarity, which is calculated by scaling the Cosine Similarity score between $-1$ and $1$ using the minimum and the maximum Cosine Similarity scores from all the experts in all the MHSA blocks of the ViT. This normalization is necessary to increase the difference in magnitudes of the similarities between tasks, which results in better Expert sampling distributions during the sampling process in our experiments. The task similarity score will be used in sampling the {\tt Reuse} and {\tt Adapt} operations.

For the new task $t$, we also have the {\tt New} expert and the {\tt Skip} expert at each layer $l$, for which we do not have similarity scores. Instead, we introduce an auxiliary expert, {\tt Aux} (see the bottom of Fig.~\ref{fig:nas-sampling}) which gives equally-likely chance to select the {\tt New} expert or the {\tt Skip} expert  once sampled in NAS. For the {\tt Aux} expert itself, the similarity score between it and the new task $t$ is specified by,
\begin{equation}
    S_l^{aux, t}=- \max_{i=1}^{t-1} S_l^{i,t} ,
\end{equation}
which intuitively means we probabilistically resort to the {\tt New} operation or the {\tt Skip} operation when other experts turn out not ``helpful'' for the task $t$.

At each layer $l$ in the ViT, for a new task $t$, the task-similarity oriented operation sampling is realized by a 2-level hierarchical sampling (Fig.~\ref{fig:nas-sampling} bottom): 
\begin{itemize} [leftmargin=*,noitemsep,topsep=0pt]
\itemsep0em
    \item The first level uses a categorical distribution with the maximum number of entries being $t$ consisting of at most the previous $t-1$ tasks (some of which may use {\tt Skip} and thus will be ignored) and the {\tt Aux} expert.  The categorical distribution $(\psi_1, \cdots, \psi_i, \cdots, \psi_{I-1}, \psi_I)$ is computed by the Softmax function over the similarity scores defined above, where $I\leq t$.  
    \item With a previous task $i$ sampled with the probability $\psi_i$, at the second level of sampling, we sample the {\tt Reuse} operation for the associated expert using a Bernoulli distribution with the success rate computed by the Sigmoid function of the task similarity score defined by $\rho_i = \frac{1}{1+\exp(-S^{i,t}_l)}$, and the {\tt Adapt} operation with probability $1-\rho_i$. 
\end{itemize}

\subsection{Compute-Aware Target Network Selection}
\label{sec:target-network-selection}

After the Supernet is trained, we propose a compute-sensitive evolutionary search on top of~\citep{real2019regularized}.
It first draws a population with a predefined number of candidate architectures from the trained Supernet using our proposed HEE sampling method. 
It then evolves the population via the crossover and the mutation operations. 
At each evolution iteration, the population is evaluated and sorted based on the trade-off between the validation performance and the compute of candidates: we predefine a performance tolerance threshold $\tau$ (e.g., $\tau=2\%$) to group candidate networks, and rank candidate networks in each group based on their compute in the increasing order. 
With the top-$k$ candidates after evaluation and sorting (the number $k$ is predefined), for crossover, two randomly sampled candidate networks in the top-$k$ are crossed to produce a new target network. For mutation, a randomly selected candidate in the top-$k$ mutates its every choice block with probability (e.g., $0.1$) to produce a new candidate. Crossover and mutation are repeated to generate sufficient new candidate target networks to form the population for the next iteration. We study the effect of varying the $\tau$ in Fig. \ref{fig:tolerance-ablation} in the supplementary. The final target network is retrained by randomly initializing all the learnable parameters following observations in~\citep{liu2018rethinking}.

\subsection{Balancing Exploration and Exploitation}\label{sec:balancing}
As illustrated in Fig.~\ref{fig:nas-sampling}, to harness the best of the pure exploration strategy and the proposed exploitation strategy, we apply epoch-wise exploration and exploitation sampling for simplicity. For the pure exploration, we directly uniformly sample the experts at a layer $l$, consisting of the $n$ experts from the previous $t-1$ tasks, and the {\tt New} and {\tt Skip} operations, where $n\leq t-1$. 
At the beginning of an epoch in the Supernet training, we choose the pure exploration strategy with a probability of $\epsilon_1$ (e.g., 0.3), and the hierarchical sampling strategy with  a probability of $1-\epsilon_1$. Similarly, when generating the initial population during the evolutionary search, we draw a candidate target network from a uniform distribution over the operations with a probability of $\epsilon_2$, and from the hierarchical sampling process with a probability of $1-\epsilon_2$, respectively. 
In practice, we set $\epsilon_2 > \epsilon_1$ (e.g., $\epsilon_2=0.5$) to encourage more exploration during the evolutionary search, while encouraging more exploitation for faster learning in the Supernet training. We study the effect of $\epsilon_1$ and $\epsilon_2$ in Fig. \ref{fig:eps-ablation} in the supplementary.

\section{Experiments}
\textbf{Data.} We evaluate CHEEM on two challenging benchmarks, the MTIL benchmark \citep{mtil} and the VDD benchmark \citep{vdd}, both consisting of tasks from varying domains with different complexities. 
VDD presents a significantly large class imbalance. For example, out of the total 2128 classes (excluding ImageNet-1k), Omniglot contains 1623 classes, whereas DTD contains only 47. Further details of the benchmarks can be found in  the supplementary (Sec.~\ref{sec:training-and-data}).

\textbf{Metrics.} We measure the performance of CHEEM using three metrics: \textbf{Average Accuracy, Average Forgetting} \citep{riemannian-walk}, and our proposed Figure-of-Merit \textbf{(FoM quantifying how many times a method is better than  the other)}. Let $(\mathcal{F}_i, \mathcal{H}_i)$ be the feature backbone and the classifier heads after completion of task $i$, and $a_{i,j} = Acc(D_j^{test}; \mathcal{F}_i, \mathcal{H}_i)$ be the Top-1 accuracy on the testing data for task $j$ computed using $(\mathcal{F}_i, \mathcal{H}_i)$. The Average Accuracy ($A\mathbb{A}$) and Average Forgetting ($A\mathbb{F}$) are respectively defined as,
\begin{align}
    A\mathbb{A} &= \frac{1}{N-1}\sum_{t=2}^N \text{Acc}(D_t^{test}; \mathcal{F}_N, \mathcal{H}_N), \label{eq:avg_accuracy_lll} \\
    A\mathbb{F} &= \frac{1}{N-2}\sum_{t=2}^{N-1} \left(\max_{j\in [t, N]} a_{j,t} - a_{N,t}\right).
    \label{eq:avg_forgetting}
\end{align}

\noindent \textbf{FoM} explicitly and holistically compares two methods (e.g., our CHEEM against another baseline) with respect to their respective average accuracies and model complexities, where the model complexity is measured using FLOPs. For two methods $m$ and $n$, we define the FoM as
\begin{equation}
    \text{FoM}(m, n) = \frac{A\mathbb{A}^{\text{UpperBound}}-A\mathbb{A}^n}{A\mathbb{A}^{\text{UpperBound}}-A\mathbb{A}^m}\cdot \frac{\text{FLOPs}^n}{\text{FLOPs}^m}, \label{eq:fom}
\end{equation}
where $A\mathbb{A}^{\text{UpperBound}}$ represents the average accuracy of upper-bound full task-to-task fine-tuning, and $\text{FLOPs}$ is the computing cost. If a method $m$ has smaller performance gap against the upper bound and smaller computing cost than another method $n$, $\text{FoM}(m,n)$ will be greater than 1. There is a trade-off between the first performance ratio and the second cost ratio. Intuitively, $\text{FoM}(m, n)$ represents the relative magnitude of method $m$ being better than $n$. 

\noindent \textbf{Pretrained Models in ExfCCL.} We test two settings: one strong \textbf{ViT-Base} pretrained on the ImageNet-21k and fine-tuned on the ImageNet-1k, and the other relatively weaker \textbf{DEiT-Tiny} trained on ImageNet-1k. We report results of \textbf{CLIP ViT-Base}~\cite{clip} in the supplementary (Sec.~\ref{sec:clip_results}).  

\textbf{Implementation Details}. In all results reported in main text, we apply CHEEM to the MLP$^{\text{Down}}$ layer, unless stated otherwise. We provide further implementation details in the supplementary (Sec.~\ref{sec:training-and-data}).

\begin{table}[t]
            \centering
            \caption{\textbf{FoM} of CHEEM (MLP$^{\text{Down}}$) against baselines on \textbf{MTIL} and \textbf{VDD} benchmarks.}
            \resizebox{0.9\linewidth}{!}{
            \begin{tabular}{r|c|c|c|c}
            \toprule
            & \multicolumn{2}{c|}{\textbf{MTIL}} & \multicolumn{2}{c}{\textbf{VDD}} \\
            \cmidrule{2-5}
            
                \textbf{Method} & \cellcolor{vitb}\textbf{ViT-Base} & \cellcolor{deit}\textbf{DEiT-Tiny} & \cellcolor{vitb}\textbf{ViT-Base} & \cellcolor{deit}\textbf{DEiT-Tiny} \\
                \midrule
                EWC  & 10.5 & 26.0 & 12.1 & 678.9 \\ 
                \hline
                CODA-P & 24.2 & 84.6 & 35.9 & 2811.3 \\
                DualP & 27.4 & 73.7 & 34.1 & 2130.6 \\
                L2P & 31.1 & 79.8 & 36.4 & 2431.6 \\
                S-Prompts & 3.2 & 10.5 & 5.5 & 338.8 \\
                DIKI & 3.6 & 6.4 & 7.8 & 370.8 \\ 
                \hline
                LoRA-CL & 1.7 & 5.7 & 1.5 & 73.6 \\ 
                Tuna & 50.6 & 242.4 & 58.6 & 4538.8 \\
                Moal & 2.38 & 7.64 & 8.63 & 462.8 \\
            \bottomrule
            \end{tabular}
            }
            \label{tab:fom-mtil}
\end{table}

\noindent \textbf{Baselines and Upper-Bounds} include: 
\begin{itemize}[leftmargin=*,noitemsep,topsep=0pt]
    \item \textit{Elastic Weight Consolidation} (EWC)~\citep{ewc}. 
    \item \textit{State-of-the-art prompt-based methods}: CODA-Prompt~\citep{coda-prompt}, Dual-Prompt~\citep{dualprompt}, Learning-to-Prompt (L2P)~\citep{learning-to-prompt}, S-Prompts~\citep{s-prompts} and DIKI~\citep{diki}.
    \item \textit{Parameter-Efficient Fine-Tuning (PEFT) based continual learning}: LoRA~\citep{lora} trained in a continual-learning setting, serving as an alternative internal-parameter memory applied to MLP$^{down}$ layer. This corresponds to a special case of CHEEM that uses the LoRA {\tt Adapt} operator at every layer and omits NAS. We refer to this setting as LoRA-CL. We further compare with two recent methods based on PEFT - Moal \cite{moal} and Tuna \cite{tuna}.
    \item \textit{Upper-bounds via task-to-task fine-tuning}: full task-to-task fine-tuning (UpperBound$_\text{Full-FT}$) and LoRA-based task-to-task PEFT (UpperBound$_\text{LoRA-FT}$), where the pretrained model is independently fine-tuned on each task to provide an upper bound on continual-learning performance.
\end{itemize}

\subsection{FoM of CHEEM Against Baselines}
Table~\ref{tab:fom-mtil} shows the FoM of CHEEM (how many times ``better" CHEEM is) on MTIL and VDD benchmarks. \textbf{The consistently significant FoM shows that CHEEM can balance Average Accuracy and FLOPs}, whereas the baseline methods fall short on one or both of the axes. 
The special case of CHEEM, LoRA-CL is close to CHEEM in terms of FoM (1.7 on MTIL and 1.5 on VDD) when ViT-Base is used, but is significantly worse when DEiT-Tiny is used (5.7 on MTIL and 73.6 on VDD). 

\subsection{CHEEM vs Upper-Bound T2T Fine-Tuning}
Table~\ref{tab:mtil-upperbound} and Table~\ref{tab:vdd-upperbound} show the 
comparisons between our CHEEM and the two upper-bound fine-tuning methods. \textbf{Our CHEEM closely approaches full fine-tuning performance}. On MTIL, CHEEM achieves 85.9\% vs 88.1\% for ViT-Base, and 74.5\% vs 75.3\% for DEiT-Tiny. On VDD, CHEEM achieves 86.7\% vs 88.7\% for ViT-Base, and 76.2\% vs 76.2\% for DEiT-Tiny. We note that FLOPs of CHEEM are nearly doubled since the task ID inference uses an additional forward computation of the initial backbones. The same is for other prompting-based methods.

\subsection{Break-Down Comparisons with Baselines}
Table~\ref{tab:mtil-agnostic-results-summary} and Table~\ref{tab:vdd-agnostic-results-summary} show the results. 
Although EWC uses the least FLOPs, it severely suffers from catastrophic forgetting due to the restriction of maintaining a single shared backbone, and can only reach average accuracy 44.58\% for ViT-Base and 35.33\% for DEiT-Tiny. 
Similarly, on both MTIL and VDD, DIKI achieves lower FLOPs, but sacrifices performance. 
The special case of CHEEM, LoRA-CL achieves Average Accuracy close to CHEEM, but requires higher FLOPs as it cannot skip modules. 
The FoM of CHEEM against  baselines for DEiT-Tiny  are significantly large on VDD (Table~\ref{tab:fom-mtil}) since CHEEM almost reaches the full fine-tuning performance (76.18\%  vs 76.21\%), resulting in a very large accuracy gap ratio term in Eqn.~\ref{eq:fom}. 

\begin{table}[t]
            \centering
            \caption{\textbf{CHEEM vs Upper-Bounds  
            on MTIL} with three seeds.}
            \vspace{2mm}
            \label{tab:mtil-upperbound}
            \resizebox{\linewidth}{!}{
            \begin{tabular}{r|c|c|c|c|c|c}
                \toprule
               \multirow{2}{*}{\textbf{Method}}  & \multicolumn{3}{c|}{\cellcolor{vitb}\textbf{ViT-Base}} & \multicolumn{3}{c}{\cellcolor{deit}\textbf{DEiT-Tiny}} \\  \cmidrule{2-7}
               &  \textbf{Avg. Acc} & \textbf{Avg. Frgt.} & \textbf{FLOPs} & \textbf{Avg. Acc} & \textbf{Avg. Frgt.} & \textbf{FLOPs} \\
               \toprule
               UpperBound$_\text{Full-FT}$  &  \textbf{88.1} $\pm$ 0.0 & - &33.7 & \textbf{75.3} $\pm$ 0.1 & - & 2.2\\
               UpperBound$_\text{LoRA-FT}$  & 87.4 $\pm$ 0.0 & - &33.7& 74.6 $\pm$ 0.1 & - & 2.2\\
               \midrule
               
               \cellcolor{highlight}CHEEM  & \cellcolor{highlight}{85.9} $\pm$ 0.3 & \cellcolor{highlight}1.7 $\pm$ 0.1 & \cellcolor{highlight}62.3 & \cellcolor{highlight}{74.5} $\pm$ 0.3 & \cellcolor{highlight}1.9 $\pm$ 0.0 & \cellcolor{highlight}4.5 \\
               \bottomrule
            \end{tabular}}
\end{table}

\begin{table}[t]
            \vspace{1mm}
            \centering
            \caption{\textbf{CHEEM vs Upper-Bounds  
            on VDD} with three seeds.}
            \vspace{2mm}
            \label{tab:vdd-upperbound}
            \resizebox{\linewidth}{!}{
            \begin{tabular}{r|c|c|c|c|c|c}
                \toprule
               \multirow{2}{*}{\textbf{Method}}  & \multicolumn{3}{c|}{\cellcolor{vitb}\textbf{ViT-Base}} & \multicolumn{3}{c}{\cellcolor{deit}\textbf{DEiT-Tiny}} \\  \cmidrule{2-7}
               &  \textbf{Avg. Acc} & \textbf{Avg. Frgt.} & \textbf{FLOPs} & \textbf{Avg. Acc} & \textbf{Avg. Frgt.} & \textbf{FLOPs} \\
               \toprule
               UpperBound$_\text{Full-FT}$  &  \textbf{88.7} $\pm$ 0.1 & - & 33.7& 76.21 $\pm$ 0.1 & - & 2.2\\
               UpperBound$_\text{LoRA-FT}$  & 86.8 $\pm$ 0.1 & - & 33.7& \textbf{76.30} $\pm$ 0.3 & - &2.2 \\
               \midrule
               
               \cellcolor{highlight}CHEEM  & \cellcolor{highlight}{86.7} $\pm$ 0.2 & \cellcolor{highlight}0.4 $\pm$ 0.0 & \cellcolor{highlight}61.6 & \cellcolor{highlight}{76.18 } $\pm$ 0.1 & \cellcolor{highlight}1.0 $\pm$ 0.0 & \cellcolor{highlight}4.5 \\               
               \bottomrule
            \end{tabular}}
\end{table}

\begin{table}[h]
            \centering
            \caption{Comparison of Average Accuracy and Forgetting  Rate
            \textbf{on the MTIL benchmark} with three seeds. 
            }
            \label{tab:mtil-agnostic-results-summary}
            \resizebox{\linewidth}{!}{
            \begin{tabular}{r|c|c|c|c|c|c}
                \toprule
               \multirow{2}{*}{\textbf{Method}}  & \multicolumn{3}{c|}{\cellcolor{vitb}\textbf{ViT-Base}} & \multicolumn{3}{c}{\cellcolor{deit}\textbf{DEiT-Tiny}} \\  \cmidrule{2-7}
               &  \textbf{Avg. Acc} & \textbf{Avg. Frgt.} & \textbf{FLOPs} & \textbf{Avg. Acc} & \textbf{Avg. Frgt.} & \textbf{FLOPs} \\
               \toprule
                             
               EWC & 44.6 $\pm$ 6.4 & 23.8 $\pm$ 6.5 & \textbf{33.7} & 35.3 $\pm$ 0.3 & 7.3 $\pm$ 0.6 & \textbf{2.2} \\ \hline
               CODA-P  & 40.2 $\pm$ 1.2 & 25.3 $\pm$ 1.8 & 70.3 & 5.6 $\pm$ 0.3 & 42.6 $\pm$ 0.8 & 5.0 \\
               DualP  & 33.8 $\pm$ 0.4 & 22.1 $\pm$ 0.4 & 70.3 & 30.9 $\pm$ 0.3 & 17.5 $\pm$ 0.3 & 5.0 \\
               L2P  & 26.6 $\pm$ 0.2 & 31.0 $\pm$ 0.3 &  70.3 & 23.2 $\pm$ 0.1 & 25.8 $\pm$ 0.4 & 5.1 \\
               S-Prompts  & 81.6 $\pm$ 0.4 & \textbf{1.6} $\pm$ 0.1 & 67.6 & 67.3 $\pm$ 0.4 & \textbf{1.8} $\pm$ 0.0 & 4.4 \\
               DIKI  & 76.4 $\pm$ 0.0 & 2.0 $\pm$ 0.0 & 42.5 & 67.6 $\pm$ 0.1 & \textbf{1.8} $\pm$ 0.0 & 2.8 \\  \hline
               LoRA-CL & 84.7 $\pm$ 0.0 & \textbf{1.6} $\pm$ 0.1 & 68.2 &  71.1 $\pm$ 0.0 & 1.9 $\pm$ 0.0 & 4.5 \\ 
               Tuna  & 70.7 $\pm$ 0.1 & 17.4 $\pm$ 0.2 & 405.0 & 44.5 $\pm$ 0.5 & 30.2 $\pm$ 0.3 & 26.0 \\  
               MoAL  & 78.3 $\pm$ 0.1 & 3.7 $\pm$ 0.1 & 33.7 & 63.8 $\pm$ 0.0 & 2.8 $\pm$ 0.0 & 2.2 \\
               
               \midrule
               \cellcolor{highlight}CHEEM  & \cellcolor{highlight}\textbf{85.9} $\pm$ 0.3 & \cellcolor{highlight}1.7 $\pm$ 0.1 & \cellcolor{highlight}62.3 & \cellcolor{highlight}\textbf{74.5} $\pm$ 0.3 & \cellcolor{highlight}1.9 $\pm$ 0.0 & \cellcolor{highlight}4.5 \\
               \bottomrule
            \end{tabular}}
\end{table}
\begin{table}[h]

            \centering
            \caption{Comparison of Average Accuracy and Forgetting Rate 
            \textbf{on the VDD benchmark} with three seeds. 
            }
            \label{tab:vdd-agnostic-results-summary}
            \resizebox{\linewidth}{!}{
            \begin{tabular}{r|c|c|c|c|c|c}
                \toprule
               \multirow{2}{*}{\textbf{Method}}  & \multicolumn{3}{c|}{\cellcolor{vitb}\textbf{ViT-Base}} & \multicolumn{3}{c}{\cellcolor{deit}\textbf{DEiT-Tiny}} \\  \cmidrule{2-7}
               &  \textbf{Avg. Acc} & \textbf{Avg. Frgt.} & \textbf{FLOPs} & \textbf{Avg. Acc} & \textbf{Avg. Frgt.} & \textbf{FLOPs} \\
               \toprule
               EWC & 44.0 $\pm$ 1.3 & 5.1 $\pm$ 1.1 & \textbf{33.7} & {33.7} $\pm$ 0.2 & 1.5 $\pm$ 0.1 & \textbf{2.2} \\ \hline
               CODA-P  & 24.9 $\pm$ 2.2 & 26.1 $\pm$ 0.8 & 70.3 & 1.1 $\pm$ 0.1 & 37.6 $\pm$ 0.4 & 5.0 \\
               DualP  & 28.1 $\pm$ 0.9 & 3.2 $\pm$ 0.5 & 70.3 & 19.4 $\pm$ 0.6 & 10.5 $\pm$ 0.5 & 5.0 \\
               L2P  & 23.9 $\pm$ 0.7 & 9.0 $\pm$ 0.6 &  70.3 & 11.5 $\pm$ 0.8 & 20.9 $\pm$ 1.7 & 5.1 \\
               S-Prompts  & 78.6 $\pm$ 0.1 & 0.4 $\pm$ 0.0 & 67.6 & 65.8 $\pm$ 0.3 & 0.9 $\pm$ 0.0 & 4.4 \\
               DIKI  & 65.9 $\pm$ 0.1 & \textbf{0.1} $\pm$ 0.0 & 42.5 & 58.3 $\pm$ 0.1 & \textbf{0.6} $\pm$ 0.0 & 2.8 \\   \hline
               LoRA-CL & 86.0 $\pm$ 0.1 & 0.3 $\pm$ 0.0 & 68.2 &  74.0 $\pm$ 0.3 & 1.1 $\pm$ 0.0 & 4.5 \\ 
               Tuna  & 70.7 $\pm$ 0.1 & 14.9 $\pm$ 0.1 & 337.0 & 47.8 $\pm$ 0.2 & 22.9 $\pm$ 0.3 & 21.5 \\  
               MoAL  & 56.7 $\pm$ 0.1 & 1.0 $\pm$ 0.0 & 33.7 & 47.9 $\pm$ 0.1 & 2.1 $\pm$ 0.0 & 2.2 \\  
                              
               \midrule
               \cellcolor{highlight}CHEEM  & \cellcolor{highlight}\textbf{86.7} $\pm$ 0.2 & \cellcolor{highlight}0.4 $\pm$ 0.0 & \cellcolor{highlight}61.6 & \cellcolor{highlight}\textbf{76.18 } $\pm$ 0.1 & \cellcolor{highlight}1.0 $\pm$ 0.0 & \cellcolor{highlight}4.5 \\
               
               \bottomrule
            \end{tabular}}
\end{table}

\subsection{Importance of Designs in CHEEM}
\label{sec:performance}

\begin{table}[t]
    \centering
    \caption{Task-wise FLOPs of CHEEM on MTIL  using ViT-Base.}
    \resizebox{0.95\linewidth}{!}{
    \begin{tabular}{llllll}
        \toprule
        \textbf{Airc} & \textbf{C101} & \textbf{CIFAR} & \textbf{DTD} & \textbf{ESAT} & \textbf{Flwr} \\ 
        62.8 $\pm$ 0.9 & 59.0 $\pm$ 0.9 & 64.0 $\pm$ 0.0 & 64.6 $\pm$ 0.9 & 57.8 $\pm$ 0.9 & 62.1 $\pm$ 0.1 \\
        \midrule
        \textbf{F101} & \textbf{MNIST} & \textbf{Pets} & \textbf{Cars} & \textbf{SUN397} & \textbf{Avg} \\
        65.2 $\pm$ 1.7 & 56.0 $\pm$ 0.9 & 62.8 $\pm$ 0.9 & 65.9 $\pm$ 1.5 & 65.2 $\pm$ 0.9 & 62.3 $\pm$ 0.3 \\ 
        \bottomrule
    \end{tabular}
    }
    \label{tab:taskwise-flops}
\end{table}

\noindent\textbf{Importance of structure updates to backbone - CHEEM vs Prompting-based Baselines}. Three prompting-based methods (CODA-Prompt, DualPrompt and L2P) perform even worse than EWC for both ViT-Base and DEiT-Tiny, mainly due to the discrepancy between global argmax vs. local argmax in their head classifier designs. CODA-Prompt almost completely fails for DEiT-Tiny with 5.62\% average accuracy. S-Prompts works the best among prompting based methods, but is still inferior to our CHEEM:  4\% drop for ViT-Base, and 7\% drop for DEiT-Tiny. This shows the importance of inferring task IDs on the fly for streaming tasks with significantly varying distributions of classes. 
Overall, \textit{the superior performance of CHEEM shows the importance of structurally and dynamically updating the backbone with the task-synergy internal memory}. 

\noindent\textbf{Importance of Search - CHEEM vs LoRA-CL}. Both are applied to MLP$^{\text{Down}}$ and use the same external task-centroid memory for task IDs inference. 
The improvement by CHEEM, 1\% increase for ViT-Base and 3.45\% increase for DEiT-Tiny show the benefits of HEE-NAS, \textit{especially for weaker backbones such as DEiT-Tiny, leading to more competent ExfCCL that is less sensitive to start backbone}. 

\subsection{CHEEM Is Unique for Task-Difficulty Awareness Using ViTs in ExfCCL}
\label{sec:task-dependent-structures}
Intuitively, easier tasks should require lesser FLOPs in continual learning. Table \ref{tab:taskwise-flops} shows that CHEEM allocates lower FLOPs to easier tasks like MNIST and ESAT. Figs.~\ref{fig:teaser-vit-b} and~\ref{fig:teaser-deit-t} show some examples of architectures learned by CHEEM on the MTIL benchmark. 
\textbf{These sensible model structures are unique to our CHEEM in comparisons to other baselines.} They also show interesting yet ``irregular'' model configurations caused by learned {\tt Skip} operations in different blocks in Fig.~\ref{fig:teaser-vit-b}: two consecutive Transformer blocks with one block comprising only the attention component (for token mixing) without the FFN (for channel mixing). Fig.~\ref{fig:teaser-vdd} in the the supplementary shows the sensible model structures learned by CHEEM on VDD.

\subsection{Reasonable Training Overhead of CHEEM}
The NAS overhead in CHEEM leads to marginal increase in total training time relative to L2P, DualPrompt, or CODA-Prompt. While it is more expensive than SPrompts, DIKI, and LoRA-CL, CHEEM lowers inference FLOPs compared to SPrompts and LoRA-CL and achieves higher average accuracy (Tables \ref{tab:mtil-agnostic-results-summary} and \ref{tab:vdd-agnostic-results-summary}). As shown in Table \ref{tab:training-time}, the full CHEEM pipeline (supernet training, search, and finetuning) remains competitive with baseline methods, offering a practical approach for learning task-specific backbones.

\begin{table}[t]
    \centering
    \caption{Average training time per task (in hours) on MTIL. The total training time of CHEEM (1.45h) is split as: Supernet - 0.6h, Search - 0.53h, Finetuning - 0.32h.}
    \resizebox{\linewidth}{!}{
    \begin{tabular}{c|c|c|c|c|c|c}
        \toprule
        \textbf{CHEEM} & \textbf{CODA-P} & \textbf{L2P} & \textbf{Dual-P} & \textbf{SPrompts} & \textbf{DIKI} & \textbf{LoRA-CL} \\
        \toprule
        1.45 & 1.26 & 1.26 & 1.25 & 0.38 & 0.38 & 0.38 \\
        \bottomrule
    \end{tabular}}
    \label{tab:training-time}\vspace{-2mm}
\end{table}

\subsection{Ablation Studies}
\label{sec:ablation-studies}

\textbf{\textbullet\, CHEEM$_{\text{lite}}$ using lightweight models for task ID learning in the external memory}: CHEEM can use a lightweight pretrained backbone for task ID learning, rather than the backbone in adaptation (Section \ref{sec:task-id-small-model}), effectively making the FLOPs overhead negligible while outperforming the baselines.

\noindent \textbf{\textbullet\, CHEEM placement: MLP$^\text{Down}$ vs. Projection }.
Table~\ref{tab:attn-vs-mlp} shows the comparisons. Both CHEEM achieve on-par average accuracy performance.  However, due to the size of the FFN block, when skipping the FFN block rather than the MHSA block, CHEEM (MLP$^\text{Down}$) shows better FLOPs reduction. 
\begin{table}[H]
    \centering
        \caption{Comparisons of two selected CHEEM placements.}
        \resizebox{1.0\linewidth}{!}{
        \begin{tabular}{r|r|c|c|c|c|c|c}
            \toprule
           \multirow{2}{*}{\textbf{Dataset}}  & \multirow{2}{*}{\textbf{Method}}  & \multicolumn{3}{c|}{\cellcolor{vitb}\textbf{ViT-Base}} & \multicolumn{3}{c}{\cellcolor{deit}\textbf{DEiT-Tiny}} \\  \cmidrule{3-8}
           & &  \textbf{Avg. Acc} & \textbf{FLOPs} & \textbf{\%Param} & \textbf{Avg. Acc} & \textbf{FLOPs} & \textbf{\%Param} \\
           \toprule
           \multirow{2}{*}{MTIL} & MLP$^\text{Down}$ & \textbf{85.88} & 62.31 & 0.40 & \textbf{74.51} & 4.47 & 6.32 \\
           & Attn Proj & 85.57 & 65.91 & 0.25 & 74.39 & 4.42 & 1.79 \\
           \bottomrule
        \end{tabular}}
        
        \label{tab:attn-vs-mlp} \vspace{-2mm}
\end{table}

\noindent \textbf{\textbullet\, Sampling in NAS: HEE vs. Uniform.}
Table~\ref{tab:hee-vs-pee} shows the comparisons. Although both methods achieve on-par average accuracy, the uniform sampling method leads to much higher number of new parameter increase (due to {\tt New} and {\tt Adapt}): 5.89\% vs 0.25\% for ViT-Base (see Figure \ref{fig:vit-b-e-full} in the the supplementary), and 14.93\% vs 6.32\% for DEiT-Tiny. \textit{The promising performance of uniform sampling based NAS shows the representational power of our proposed internal parameter memory using the four basic operations ({\tt Reuse, Adapt, New} and {\tt Skip}). The parsimoniousness of HEE-NAS highlights its efficacy in continual learning by effectively leveraging task synergies.}
\begin{table}[H]
    \centering
        \caption{Comparisons of HEE and Uniform (i.e., Pure Exploration) sampling during Supernet training on MTIL benchmark.}
        \resizebox{1.0\linewidth}{!}{
        \begin{tabular}{r|c|c|c|c|c|c}
            \toprule
           \multirow{2}{*}{\textbf{Method}}  & \multicolumn{3}{c|}{\cellcolor{vitb}\textbf{ViT-Base}} & \multicolumn{3}{c}{\cellcolor{deit}\textbf{DEiT-Tiny}} \\  \cmidrule{2-7}
           &  \textbf{Avg. Acc} & \textbf{FLOPs} & \textbf{\%Param} & \textbf{Avg. Acc} & \textbf{FLOPs} & \textbf{\%Param} \\
           \toprule
           HEE  & \textbf{85.88} & 62.31 & 0.25 & {74.51} & 4.47 & 6.32 \\
           Uniform & 84.74 & 61.82 & 5.89 & \textbf{75.05} & 4.47 & 14.93 \\
           \bottomrule
        \end{tabular}}
        \label{tab:hee-vs-pee} \vspace{-2mm}
\end{table}

\noindent \textbf{\textbullet\, Effect of task orders}. In Table \ref{tab:mtil-task-order} in the supplementary, we show that CHEEM is insensitive to task order.

\section{Conclusion}
This paper presents a method of transforming Vision Transformers (ViTs) for exemplar-free class-incremental continual learning (ExfCCL), dubbed \textbf{CHEEM} (Continual Hierarchical-Exploration-Exploitation Memory).  The core of CHEEM is its internal (parameter) memory, 
which is realized by a proposed Hierarchical-Exploration-Exploitation (HEE) sampling based neural architecture search algorithm. 
CHEEM is tested on two challenging benchmarks, the MTIL and VDD benchmarks. It obtains state-of-the-art performance on both benchmarks, outperforming the prior art by a large margin, with sensible CHEEM structures continually learned.

\subsubsection*{Acknowledgments}
This work was supported in part by ARO Grants W911NF1810295 and W911NF2210010, NSF awards IIS-1909644, CMMI-2024688, and IUSE-2013451, as well as the NC State Goodnight Early Career Award. Portions of M. Dai’s contributions were completed while she was an undergraduate student at Princeton University.
The views and conclusions expressed in this paper are those of the authors and do not necessarily reflect the official policies or endorsements, either expressed or implied, of ARO, NSF, or the U.S. Government. The U.S. Government is authorized to reproduce and distribute reprints for governmental purposes notwithstanding any copyright notation herein.

%% file: sec/appendix_arxiv.tex
\section{Examples of CHEEM learned continually on the VDD benchmark}
\label{sec:additional_results}
\begin{figure*} [!htb]
    \centering
    \begin{subfigure}{0.8\textwidth}
        \centering
        \includegraphics[width=.99\linewidth]{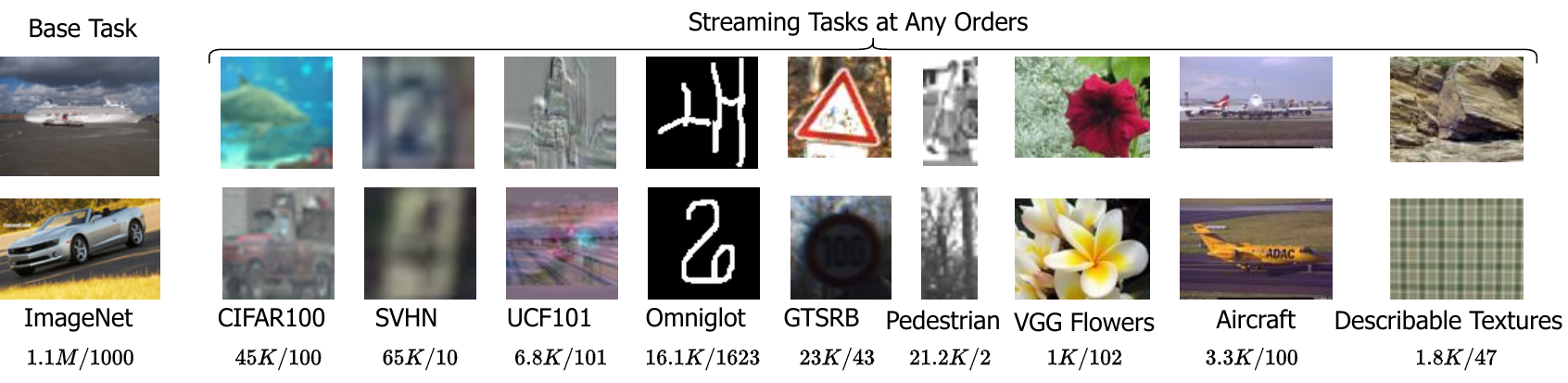}
        \caption{\textbf{The VDD benchmark}~\cite{vdd} consisting of tasks of different nature with {\tt \#training images/\#classes} significantly varying across different tasks. }
        \label{fig:vdd-teaser}
    \end{subfigure}
    \vspace{2mm}
    
    \begin{subfigure}{0.8\textwidth}
        \centering
        \includegraphics[width=.99\linewidth]{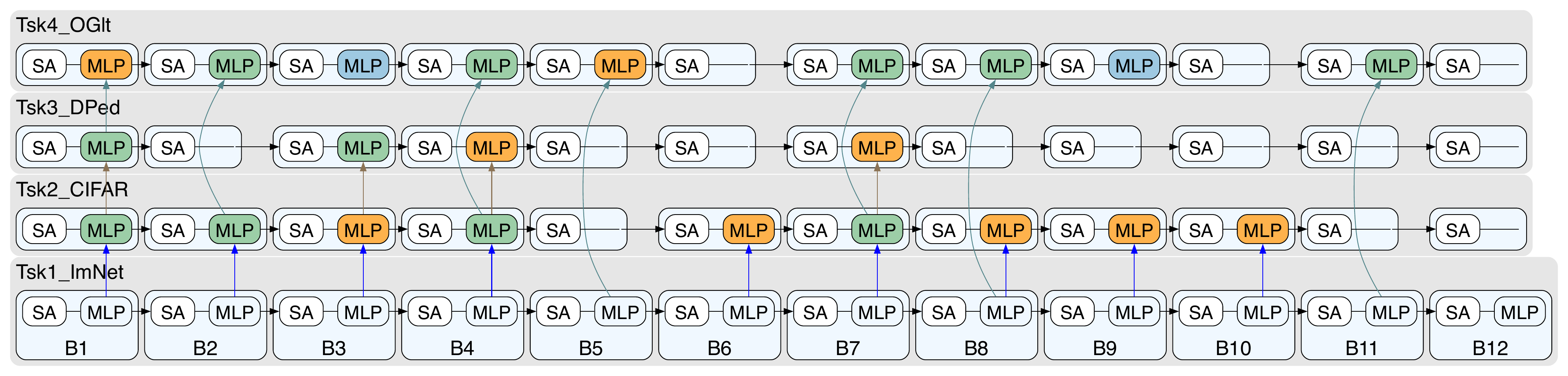}
        \caption{From \textbf{ViT-Base trained on Tsk1\_ImNet} (with blocks B1 to B12), our CHEEM learns sensible task-tailored models that reflect task complexity. For example, when learning Daimer Pedestrian Classification (Tsk3\_DPed), CHEEM learns to {\tt Skip} 8 MLP blocks and \colorbox{reuse}{Reuse} most of the architecture. When learning Omniglot (Tsk3\_Oglt), which has a larger shift from ImageNet, CHEEM learns to \colorbox{adapt}{Adapt} the ImageNet parameters in Blocks 1 and 5, adds  \colorbox{new}{New} operations in Blocks 3 and 9, and {\tt Skips} blocks 6, 10 and 12.}
        \label{fig:vdd-teaser-vit-b}
    \end{subfigure}
    \vspace{2mm}
    
    \begin{subfigure}{0.8\textwidth}
        \centering
        \includegraphics[width=0.99\linewidth]{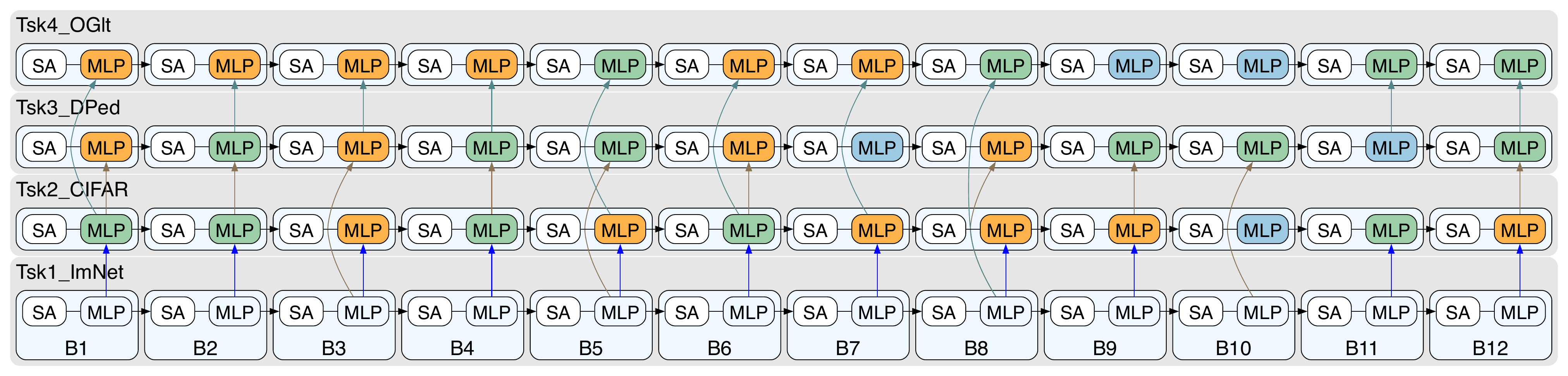}
        \caption{From \textbf{DEiT-Tiny trained on Tsk1\_ImNet} (with blocks B1 to B12), our CHEEM learns to use multiple \colorbox{adapt}{Adapt} and \colorbox{new}{New} operations, without {\tt Skip} operations selected, sensibly different from those with more {\tt Skip} and less \colorbox{new}{New} operations learned based on the stronger ViT-Base model. }
        \label{fig:vdd-teaser-deit-t}
    \end{subfigure}
    \vspace{4mm}
    
    \caption{Examples of CHEEM learning task-tailored models.}
    \label{fig:teaser-vdd}\vspace{-3mm}
\end{figure*}

\clearpage

\section{Effects of streaming task orders}
\label{sec:task-order}
We verify the effect of different task orders on the performance of CHEEM. Table \ref{tab:mtil-task-order} shows that CHEEM is robust to task orders on the MTIL benchmark.

\begin{table*}[!htb]
    \centering
    \caption{Results of learning CHEEM on the MTIL benchmark with three different streaming task orders.}
    \resizebox{0.9\textwidth}{!}{
    \begin{tabular}{ccccccccccc|c|c}
        \toprule
        \textbf{SUN} & \textbf{Airc} & \textbf{DTD} & \textbf{F101} & \textbf{Cars} & \textbf{C101} & \textbf{CIFAR} & \textbf{ESAT} & \textbf{Flwr} & \textbf{MNIST} & \textbf{Pets} & \textbf{Avg. Acc.} & \textbf{Avg. Frgt.} \\
        68.59 & 67.87 & 69.15 & 89.02 & 83.60 & 84.41 & 90.47 & 98.56 & 97.82 & 99.65 & 93.05 & 85.65 & 1.38 \\
        \midrule
        \textbf{C101} & \textbf{CIFAR} & \textbf{ESAT} & \textbf{Flwr} & \textbf{MNIST} & \textbf{Pets} & \textbf{DTD} & \textbf{Cars} & \textbf{F101} & \textbf{Airc} & \textbf{SUN} & \textbf{Avg. Acc.} & \textbf{Avg. Frgt.} \\
        78.23 & 90.36 & 98.42 & 97.76 & 99.66 & 91.77 & 69.15 & 84.48 & 89.30 & 66.13 & 66.86 & 84.74 & 2.47 \\
        \midrule
        \textbf{MNIST} & \textbf{SUN} & \textbf{Flwr} & \textbf{DTD} & \textbf{C101} & \textbf{Cars} & \textbf{Pets} & \textbf{F101} & \textbf{CIFAR} & \textbf{Airc} & \textbf{ESAT} & \textbf{Avg. Acc.} & \textbf{Avg. Frgt.} \\
        99.63 & 68.58 & 98.11 & 67.87 & 84.52 & 84.39 & 92.53 & 88.50 & 90.88 & 69.10 & 97.78 & 85.63 & 1.28 \\
        \bottomrule
        \end{tabular}}
    \label{tab:mtil-task-order}
\end{table*}

\section{Full Learned CHEEM on MTIL}
\label{sec:full_structures}
\begin{figure*}[!htb]
    \centering
    \includegraphics[width=0.99\linewidth]{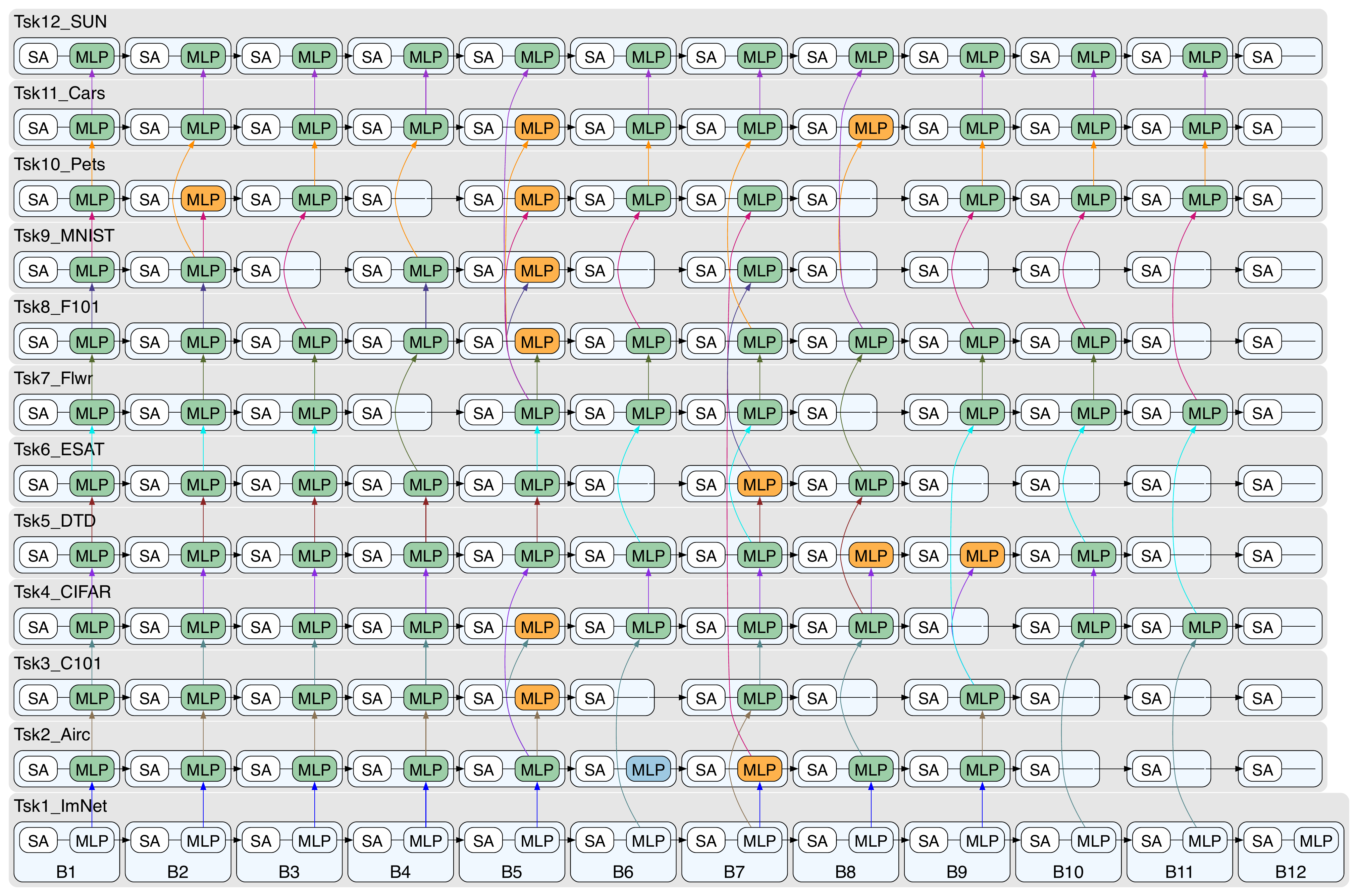}
    \caption{Figure \ref{fig:teaser-vit-b} from the main text reproduced on the full benchmark. From \textbf{ViT-Base trained on Tsk1\_ImNet} (with blocks B1 to B12), our CHEEM learns sensible task-tailored models that reflect task complexity. For example, when learning Caltech 101 (Tsk3\_C101), CHEEM learns to {\tt Skip} 5 MLP blocks and \colorbox{reuse}{Reuse} most of the architecture. In contrast, when learning FGVC Aircraft (Tsk1\_Airc), which is a more complex task with larger distribution shift from ImageNet due to its fine-grained nature, CHEEM learns to \colorbox{adapt}{Adapt} the ImageNet parameters in Block 7, adds a \colorbox{new}{New} operation in Block 6, and {\tt Skips} the last 3 MLP blocks. When learning MNIST, CHEEM {\tt skips} 8 MLP blocks, accounting for the easy nature of the task.}
    \label{fig:vit-b-ee-full}
\end{figure*}

\begin{figure*}[!htb]
    \centering
    \includegraphics[width=0.99\linewidth]{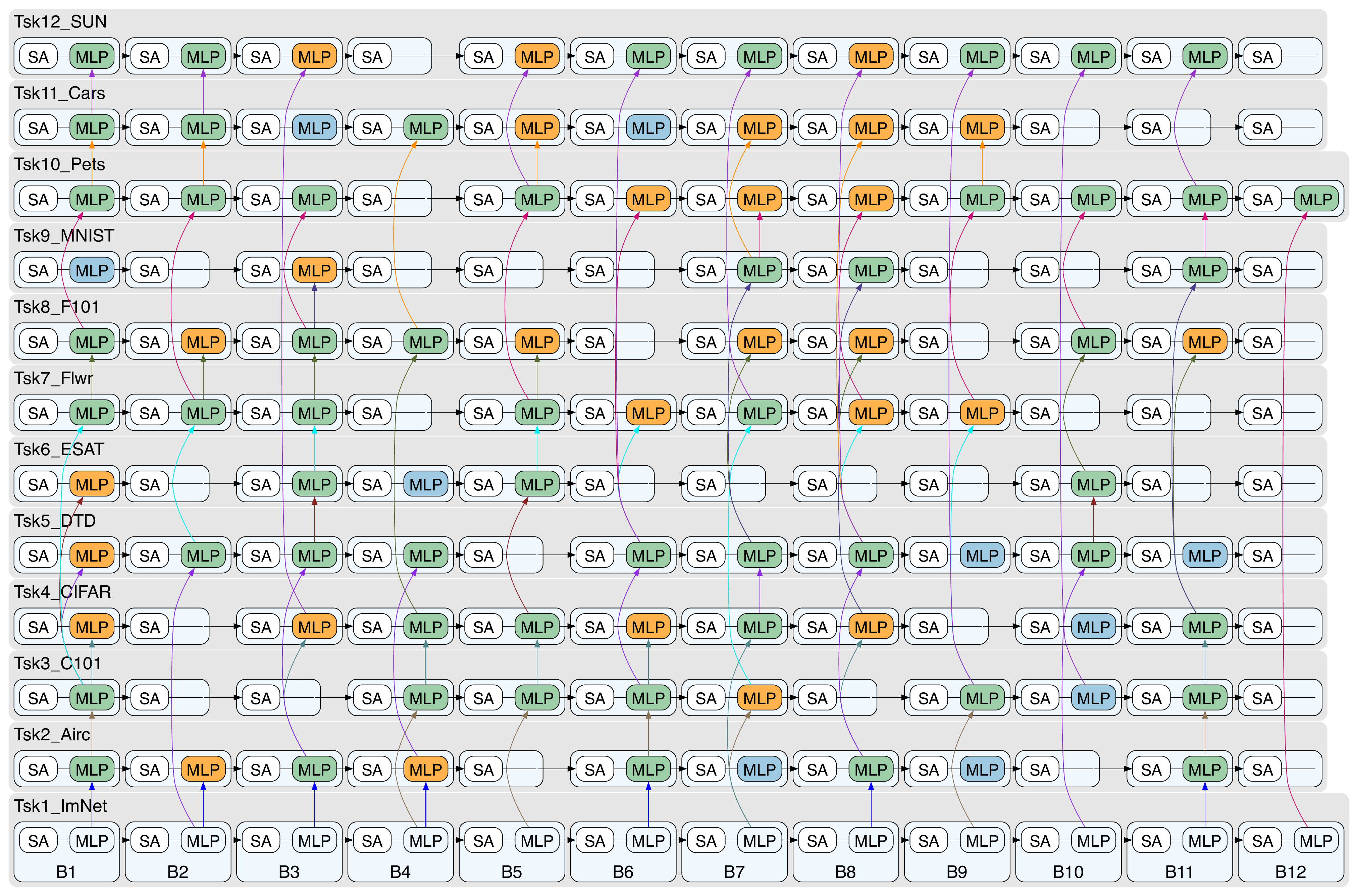}
    \caption{\textbf{ViT-Base trained on Tsk1\_ImNet} (with blocks B1 to B12), with \textbf{Pure Exploration in CHEEM}. While pure exploration accounts for task complexity through the {\tt skip} operation, it also adds more many more \colorbox{adapt}{Adapt} and \colorbox{new}{New} operations as compared to the proposed Hierarchical Exploration-Exploitation scheme (Figure \ref{fig:vit-b-ee-full}). This shows that the HEE sampling scheme can effectively leverage task synergies and reuse previous parameter memories.}
    \label{fig:vit-b-e-full}
\end{figure*}

\begin{figure*}[t]
    \centering
    \includegraphics[width=0.99\linewidth]{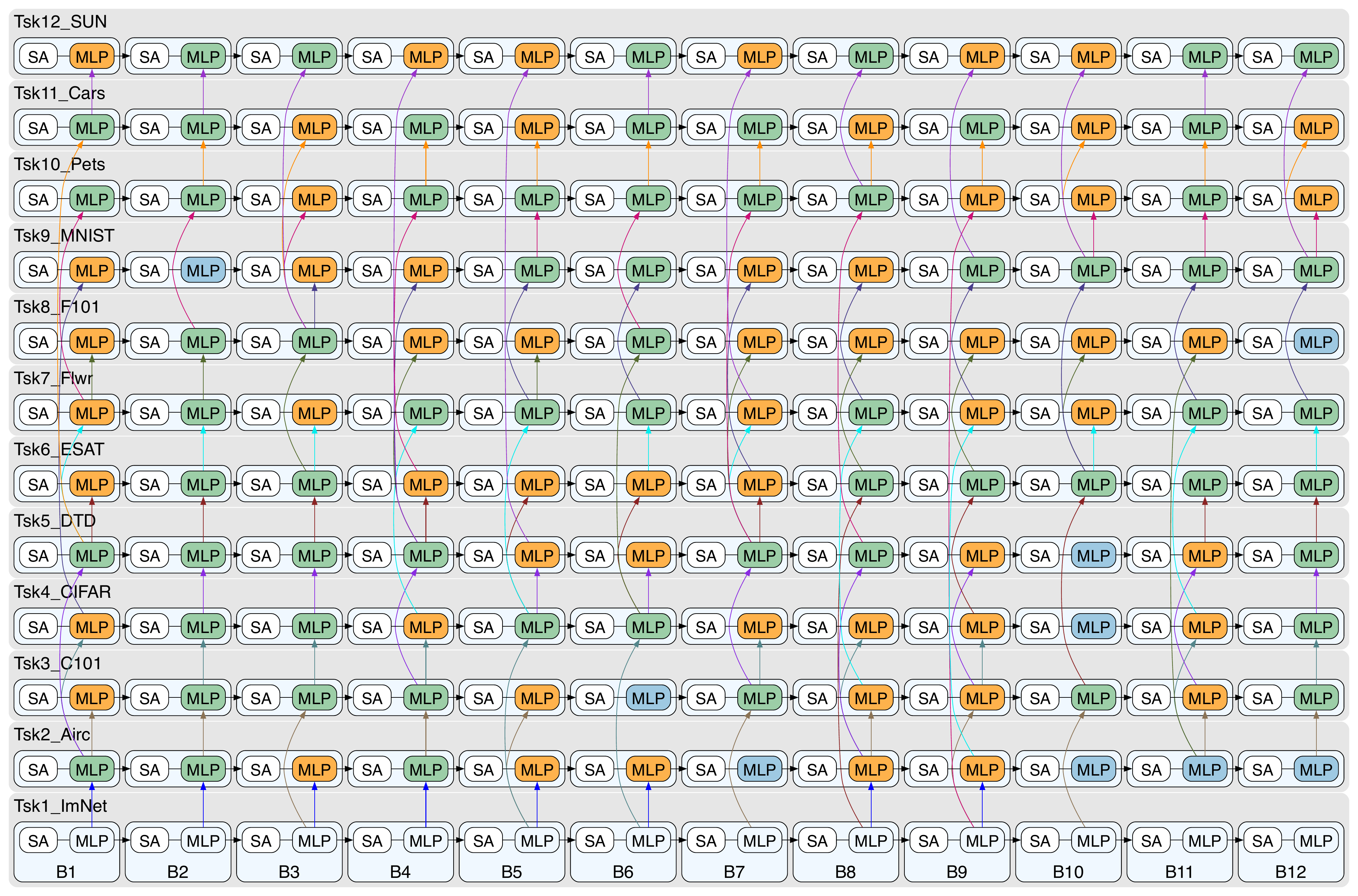}
    \caption{Figure \ref{fig:teaser-deit-t} reproduced on the full benchmark. From \textbf{DEiT-Tiny trained on Tsk1\_ImNet} (with blocks B1 to B12), our CHEEM learns to use multiple \colorbox{adapt}{Adapt} and \colorbox{new}{New} operations, without {\tt Skip} operations selected, sensibly different from those with more {\tt Skip} and less \colorbox{new}{New} operations learned based on the stronger ViT-Base model.}
    \label{fig:diet-tiny-ee}
\end{figure*}
\clearpage

\section{Full Results}
\label{sec:full_results}
\begin{table*}[!htb]
    \centering
    \caption{\textbf{MTIL}: Full results on the MTIL benchmark, extending Tables \ref{tab:mtil-upperbound} and \ref{tab:mtil-agnostic-results-summary} in the main text.}
    \resizebox{0.9\linewidth}{!}{
    \begin{tabular}{c|c|c|c|c|c|c|c|c|c|c|c|c|c}
        \toprule
       \textbf{Method}  & \textbf{Airc} & \textbf{C101} & \textbf{CIFAR} & \textbf{DTD} & \textbf{ESAT} & \textbf{Flwr} & \textbf{F101} & \textbf{MNIST} & \textbf{Pets} & \textbf{Cars} & \textbf{SUN} & \textbf{Avg. Acc} & \textbf{Avg. Frgt.} \\
       \toprule
       \multicolumn{14}{c}{\cellcolor{vitb}\textbf{ViT-Base}} \\
       \midrule
       Full Finetuning  & 69.87 & 98.32 & 90.66 & 77.46 & 98.78 & 97.87 & 88.46 & 99.70 & 92.85 & 85.42 & 69.89 & 88.12 $\pm$ 0.04 & - \\
       LoRA Finetuning  & 63.86 & 97.77 & 91.35 & 77.59 & 98.84 & 98.83 & 88.41 & 99.69 & 93.14 & 80.26 & 71.96 & 87.43 $\pm$ 0.01 & - \\
       \midrule
       CHEEM (MLP$^{Down}$, HEE)  & 69.77 & 84.86 & 90.27 & 68.48 & 98.31 & 97.54 & 89.48 & 99.60 & 92.88 & 84.94 & 68.58 & 85.88 $\pm$ 0.29 & 1.73 $\pm$ 0.05 \\
       CHEEM (MLP$^{Down}$, PE)  & 69.97 & 84.96 & 90.21 & 66.74 & 97.97 & 97.32 & 86.97 & 99.50 & 92.32 & 82.11 & 64.06 & 84.74 $\pm$ 0.26 & 1.72 $\pm$ 0.05 \\
       CHEEM (Attn Proj, HEE)  & 69.92 & 83.00 & 90.44 & 66.54 & 98.31 & 97.53 & 88.94 & 99.60 & 92.90 & 85.50 & 68.62 & 85.57 $\pm$ 0.27 & 1.67 $\pm$ 0.03 \\
       \midrule
       EWC & 39.10 & 40.90 & 43.93 & 12.98 & 61.43 & 22.24 & 51.81 & 96.20 & 60.65 & 12.64 & 48.46 & 44.58 $\pm$ 6.35 & 23.80 $\pm$ 6.53 \\
       CODA-Prompt  & 0.91 & 19.14 & 75.60 & 7.39 & 38.26 & 24.40 & 84.62 & 97.32 & 36.02 & 12.61 & 46.17 & 40.22 $\pm$ 1.22 & 25.25 $\pm$ 1.78 \\
       DualPrompt  & 3.08 & 14.40 & 83.96 & 3.48 & 46.45 & 6.00 & 85.46 & 68.43 & 24.13 & 5.88 & 30.78 & 33.82 $\pm$ 0.35 & 22.11 $\pm$ 0.42 \\
       L2P  & 1.22 & 17.35 & 78.81 & 3.46 & 30.39 & 4.67 & 78.47 & 16.83 & 23.45 & 4.62 & 33.40 & 26.61 $\pm$ 0.16 & 30.96 $\pm$ 0.27 \\
       S-Prompts  & 53.78 & 82.54 & 88.26 & 65.44 & 96.71 & 98.51 & 84.64 & 99.23 & 92.88 & 70.09 & 65.79 & 81.62 $\pm$ 0.35 & 1.64 $\pm$ 0.05 \\
       DIKI  & 52.29 & 91.68 & 89.10 & 63.95 & 96.31 & 30.22 & 86.55 & 98.37 & 92.24 & 70.22 & 69.74 & 76.42 $\pm$ 0.04 & 1.96 $\pm$ 0.02 \\
       LoRA (MLP$^{Down}$) & 63.78 & 85.68 & 90.52 & 67.98 & 98.41 & 98.51 & 87.26 & 99.69 & 92.51 & 79.79 & 67.59 & 84.70 $\pm$ 0.01 & 1.64 $\pm$ 0.11 \\
       
       \midrule
       \multicolumn{14}{c}{\cellcolor{deit}\textbf{DEiT Tiny}} \\
       \midrule
       \textbf{Method}  & \textbf{Airc} & \textbf{C101} & \textbf{CIFAR} & \textbf{DTD} & \textbf{ESAT} & \textbf{Flwr} & \textbf{F101} & \textbf{MNIST} & \textbf{Pets} & \textbf{Cars} & \textbf{SUN} & \textbf{Avg. Acc} & \textbf{Avg. Frgt.} \\
       \midrule
       Full Finetuning  & 43.17 & 94.64 & 83.55 & 64.88 & 98.67 & 68.90 & 79.88 & 99.65 & 86.57 & 54.83 & 52.99 & 75.25 $\pm$ 0.12 & - \\
       LoRA Finetuning  & 39.92 & 93.71 & 81.04 & 63.37 & 98.59 & 74.38 & 76.25 & 99.58 & 87.38 & 53.80 & 52.97 & 74.64 $\pm$ 0.08 & - \\
       \midrule
       CHEEM (MLP$^{Down}$, HEE) & 52.51 & 80.59 & 79.67 & 57.43 & 97.86 & 73.94 & 77.89 & 99.60 & 87.37 & 61.73 & 51.02 & 74.51 $\pm$ 0.28 & 1.86 $\pm$ 0.04 \\
       CHEEM (MLP$^{Down}$, PE) & 53.03 & 80.50 & 80.16 & 57.66 & 97.86 & 80.37 & 78.11 & 99.62 & 85.95 & 62.43 & 49.81 & 75.05 $\pm$ 0.12 & 1.85 $\pm$ 0.06 \\
       CHEEM (Attn Proj, HEE) & 50.65 & 80.35 & 78.44 & 56.77 & 97.75 & 77.23 & 77.71 & 99.55 & 86.84 & 61.72 & 51.27 & 74.39 $\pm$ 0.13 & 1.95 $\pm$ 0.03 \\
       \midrule
       EWC & 37.38 & 13.94 & 48.87 & 0.00 & 83.14 & 0.00 & 50.65 & 93.72 & 30.44 & 2.89 & 27.57 & 35.33 $\pm$ 0.32 & 7.34 $\pm$ 0.55 \\
       CODA-Prompt  & 0.00 & 1.77 & 2.75 & 0.04 & 0.32 & 0.00 & 22.46 & 3.94 & 5.80 & 0.27 & 24.45 & 5.62 $\pm$ 0.25 & 42.58 $\pm$ 0.81 \\
       DualPrompt  & 0.66 & 42.28 & 59.13 & 3.03 & 42.04 & 0.86 & 42.10 & 55.06 & 47.42 & 5.75 & 41.47 & 30.89 $\pm$ 0.29 & 17.53 $\pm$ 0.27 \\
       L2P & 0.11 & 39.46 & 47.87 & 4.11 & 29.80 & 1.02 & 37.07 & 0.83 & 50.15 & 1.29 & 43.97 & 23.24 $\pm$ 0.14 & 25.81 $\pm$ 0.37 \\
       S-Prompts & 36.00 & 79.08 & 71.58 & 50.50 & 93.87 & 72.27 & 67.97 & 98.66 & 87.44 & 40.01 & 43.22 & 67.33 $\pm$ 0.38 & 1.80 $\pm$ 0.02 \\
       DIKI & 33.95 & 76.57 & 71.13 & 54.84 & 92.66 & 71.79 & 70.46 & 97.40 & 87.61 & 40.13 & 47.41 & 67.63 $\pm$ 0.06 & 1.76 $\pm$ 0.01 \\
       LoRA (MLP$^{Down}$) & 39.48 & 78.89 & 78.11 & 54.38 & 97.80 & 73.66 & 74.80 & 99.58 & 85.88 & 53.53 & 45.61 & 71.06 $\pm$ 0.02 & 1.87 $\pm$ 0.00 \\
       \bottomrule
    \end{tabular}}
    \label{tab:mtil-agnostic-results-full}
\end{table*}

\begin{table*}[!htb]
    \centering
    \caption{\textbf{VDD}: Full results on VDD benchmark, extending Table \ref{tab:vdd-upperbound} and \ref{tab:vdd-agnostic-results-summary} in the main text.}
    \resizebox{0.9\linewidth}{!}{
    \begin{tabular}{c|c|c|c|c|c|c|c|c|c|c|c}
        \toprule
       \textbf{Method}  & \textbf{CIFAR} & \textbf{DPed} & \textbf{OGlt} & \textbf{SVHN} & \textbf{UCF} & \textbf{GTSR} & \textbf{Flwr} & \textbf{Airc} & \textbf{DTD} & \textbf{Avg. Acc} & \textbf{Avg. Frgt.} \\
       \toprule
       \multicolumn{12}{c}{\cellcolor{vitb}\textbf{ViT-Base}} \\
       \midrule
       Full Finetuning & 90.65 & 99.97 & 86.06 & 97.75 & 79.54 & 99.35 & 98.03 & 70.29 & 76.99 & 88.74 $\pm$ 0.11 & - \\
       LoRA Finetuning & 91.44 & 99.50 & 79.43 & 97.42 & 73.36 & 98.95 & 98.96 & 64.03 & 77.64 & 86.75 $\pm$ 0.11 & - \\
       
       \midrule
       CHEEM (MLP$^{Down}$, HEE) & 90.06 & 99.59 & 83.32 & 95.87 & 73.96 & 97.09 & 97.48 & 67.13 & 75.85 & 86.71 $\pm$ 0.23 & 0.35 $\pm$ 0.02 \\
       CHEEM (Attn Proj, HEE) & 89.90 & 99.58 & 83.08 & 96.26 & 74.49 & 97.27 & 97.56 & 70.55 & 76.42 & 87.23 $\pm$ 0.22 & 0.34 $\pm$ 0.01 \\
       \midrule
       EWC & 83.69 & 97.69 & 6.91 & 77.43 & 25.92 & 78.20 & 0.06 & 5.98 & 19.91 & 43.98 $\pm$ 1.34 & 5.09 $\pm$ 1.14 \\
       CODA-Prompt  & 37.69 & 1.29 & 6.87 & 54.52 & 2.32 & 49.22 & 39.18 & 7.48 & 25.16 & 24.86 $\pm$ 2.19 & 26.11 $\pm$ 0.75 \\
       DualPrompt  & 82.34 & 4.04 & 14.37 & 15.02 & 13.41 & 64.42 & 27.20 & 15.29 & 16.37 & 28.05 $\pm$ 0.85 & 3.18 $\pm$ 0.51 \\
       L2P  & 86.64 & 4.98 & 14.75 & 6.63 & 14.19 & 27.89 & 25.59 & 16.71 & 18.12 & 23.94 $\pm$ 0.72 & 8.98 $\pm$ 0.64 \\
       S-Prompts  & 88.34 & 99.47 & 57.38 & 94.23 & 55.07 & 87.90 & 98.48 & 53.52 & 72.59 & 78.55 $\pm$ 0.09 & 0.36 $\pm$ 0.04 \\
       DIKI  & 86.54 & 98.20 & 57.70 & 63.44 & 52.10 & 72.66 & 36.45 & 53.53 & 72.82 & 65.94 $\pm$ 0.05 & 0.11 $\pm$ 0.01 \\
       LoRA  (MLP$^{Down}$) & 90.18 & 99.21 & 79.43 & 96.35 & 73.10 & 97.39 & 98.54 & 64.01 & 76.19 & 86.04 $\pm$ 0.11 & 0.34 $\pm$ 0.03 \\
       \midrule
       \multicolumn{12}{c}{\cellcolor{deit}\textbf{DEiT Tiny}} \\
       \midrule
       \textbf{Method}  & \textbf{CIFAR} & \textbf{DPed} & \textbf{OGlt} & \textbf{SVHN} & \textbf{UCF} & \textbf{GTSR} & \textbf{Flwr} & \textbf{Airc} & \textbf{DTD} & \textbf{Avg. Acc} & \textbf{Avg. Frgt.} \\
       \midrule
       Full Finetuning & 83.50 & 99.97 & 69.71 & 97.24 & 57.97 & 98.95 & 69.04 & 44.46 & 65.02 & 76.21 $\pm$ 0.07 & - \\
       LoRA Finetuning & 81.29 & 99.96 & 76.93 & 96.37 & 54.83 & 98.16 & 74.37 & 40.66 & 63.67 & 76.25 $\pm$ 0.30 & - \\
       \midrule
       CHEEM (MLP$^{Down}$, HEE) & 75.75 & 97.73 & 81.64 & 95.30 & 57.26 & 93.11 & 74.76 & 45.91 & 64.13 & 76.18 $\pm$ 0.10 & 1.03 $\pm$ 0.01 \\
       CHEEM (Attn Proj, HEE) & 74.70 & 97.85 & 80.43 & 95.22 & 57.46 & 93.68 & 75.75 & 46.55 & 62.11 & 75.97 $\pm$ 0.36 & 1.09 $\pm$ 0.01 \\
       \midrule
       EWC & 79.39 & 93.96 & 0.03 & 60.13 & 4.97 & 64.41 & 0.00 & 0.58 & 0.00 & 33.72 $\pm$ 0.15 & 1.52 $\pm$ 0.08 \\
       CODA-Prompt  & 2.07 & 0.00 & 0.02 & 1.55 & 0.02 & 0.56 & 0.36 & 0.30 & 5.16 & 1.12 $\pm$ 0.08 & 37.56 $\pm$ 0.40 \\
       DualPrompt  & 47.87 & 4.48 & 28.60 & 11.53 & 2.54 & 75.67 & 0.40 & 0.57 & 2.61 & 19.36 $\pm$ 0.55 & 10.54 $\pm$ 0.49 \\
       L2P & 56.24 & 1.38 & 0.80 & 0.26 & 2.15 & 37.43 & 1.24 & 0.17 & 3.90 & 11.51 $\pm$ 0.76 & 20.90 $\pm$ 1.72 \\
       S-Prompts & 68.58 & 97.24 & 46.05 & 85.87 & 43.44 & 80.13 & 74.78 & 36.72 & 58.90 & 65.75 $\pm$ 0.27 & 0.90 $\pm$ 0.02 \\
       DIKI & 65.54 & 97.44 & 44.89 & 45.55 & 40.78 & 64.49 & 72.37 & 34.41 & 59.38 & 58.32 $\pm$ 0.05 & 0.62 $\pm$ 0.00 \\
       LoRA (MLP$^{Down}$) & 74.26 & 97.69 & 76.87 & 94.96 & 52.68 & 93.09 & 73.75 & 40.52 & 62.22 & 74.01 $\pm$ 0.34 & 1.07 $\pm$ 0.02 \\
       \bottomrule
    \end{tabular}}
    \label{tab:vdd-agnostic-results-full}
\end{table*}

\section{Effect of Exploration Probability ($\epsilon_1$, $\epsilon_2$) and Tolerance Threshold ($\tau$)}
\label{sec:epsilon-ablations}
\begin{figure*}[!ht]
    \centering
    \begin{subfigure}[t]{0.48\linewidth}
        \centering
        \includegraphics[width=\linewidth]{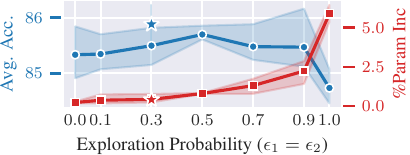}
        \newsubcap{Effect of the exploration probability on the MTIL benchmark, with exploration probabilities $\epsilon_1$ (supernet training) and $\epsilon_2$ (evolutionary search) set equal. As $\epsilon$ increases, \textcolor{tabblue}{average accuracy} first rises, then falls, while the \textcolor{tabred}{average number of additional parameters per task} increases monotonically. This is due to more new operations being learned; $\epsilon=0.3$ strikes a good balance. \textbf{Setting $\epsilon < 0.5$ controls the addition of new operations while maintaining performance}. $\epsilon_1=0.3$ and $\epsilon_2=0.5$ used in our experiments (denoted by $\bigstar$) improve accuracy further without increasing parameters. In sum, $\epsilon$ governs the number of {\tt reuse} (exploitation), {\tt adapt}, and {\tt new} (exploration) operations. 
        }
        \label{fig:eps-ablation}
    \end{subfigure}
    ~
    \begin{subfigure}[t]{0.48\linewidth}
        \centering
        \includegraphics[width=\linewidth]{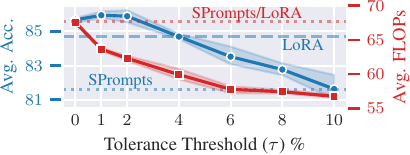}
        \newsubcap{
        A higher Tolerance Threshold reduces \textcolor{tabred}{\textbf{average FLOPs per task}} but also lowers \textcolor{tabblue}{\textbf{average accuracy}}, as it permits more {\tt skip} operations to persist in the population during evolutionary search, even if their accuracy is lower (within the tolerance margin). A 2\% threshold, used in our experiments, offers a good trade-off. At $\tau=6\%$, CHEEM still surpasses \textcolor{tabblue}{SPrompts in average accuracy (dotted blue line)} while using significantly fewer FLOPs, beyond which the FLOPs plateau. \textcolor{tabred}{SPrompt FLOPs (dotted red line)} closely match those of LoRA, so the same line is used. At $\tau=4\%$, CHEEM matches \textcolor{tabblue}{LoRA's average accuracy (dashed blue line)} with substantially fewer FLOPs. Thus, with $\tau \leq 4\%$, CHEEM matches or exceeds LoRA in accuracy while reducing FLOPs.}
        \label{fig:tolerance-ablation}
    \end{subfigure}
    \vspace{-2  mm}
\end{figure*}

\section{Generalization to non-ImageNet backbones}
\label{sec:clip_results}

Table~\ref{tab:clip} shows that CHEEM is effective on models pretrained on datasets and objectives beyond ImageNet.
\begin{table}[!h]
    \centering
    \caption{Results on the \textbf{MTIL} benchmark using \textbf{CLIP} ViT-B/16~\cite{clip}.}
    \resizebox{0.5\linewidth}{!}{
    \begin{tabular}{c|c|c|c}
        \toprule
        \textbf{Method} & \textbf{Avg. Acc.} $\uparrow$ & \textbf{Frgt} $\downarrow$ & \textbf{FLOPs} $\downarrow$ \\
        \toprule
        CODA-P & 27.6 $\pm$ 1.4 & 42.1 $\pm$ 1.4 & 70.3 \\
        L2P & 37.2 $\pm$ 0.1 & 23.9 $\pm$ 0.6 & 70.3 \\
        DualPrompt & 44.6 $\pm$ 0.4 & 15.0 $\pm$ 0.4 & 70.3 \\
        S-Prompts & 83.1 $\pm$ 0.1 & 1.2 $\pm$ 0.0 & 67.6 \\
        DIKI & 77.9 $\pm$ 0.1 & \textbf{1.0} $\pm$ 0.0 & \textbf{42.5} \\
        LoRA-CL & \textbf{86.8} $\pm$ 0.3 & 1.2 $\pm$ 0.0 & 68.2 \\
        Tuna & 66.7 $\pm$ 0.3 & 22.7 $\pm$ 0.4 & 405.0 \\
        T2T-Finetune & 86.3 $\pm$ 0.1 & - & - \\
        \midrule
        \cellcolor{highlight}Our CHEEM & \cellcolor{highlight}\textbf{86.4} $\pm$ 0.2 & \cellcolor{highlight}1.2 $\pm$ 0.0 & \cellcolor{highlight}64.3 \\
        \bottomrule
    \end{tabular}}
    \label{tab:clip}\vspace{-2mm}
\end{table}

\begin{table}[!h]
    \centering
    \caption{Results on the \textbf{VDD} benchmark using \textbf{CLIP} ViT-B/16.}
    \resizebox{0.5\linewidth}{!}{
    \begin{tabular}{c|c|c|c}
        \toprule
        \textbf{Method} & \textbf{Avg. Acc.} $\uparrow$ & \textbf{Frgt} $\downarrow$ & \textbf{FLOPs} $\downarrow$ \\
        \toprule
        CODA-P & 16.3 $\pm$ 0.7 & 34.4 $\pm$ 0.4 & 70.3 \\
        L2P & 23.0 $\pm$ 1.1 & 21.6 $\pm$ 1.2 & 70.3 \\
        DualPrompt & 29.8 $\pm$ 1.4 & 16.8 $\pm$ 1.8 & 70.3 \\
        S-Prompts & 78.4 $\pm$ 0.1 & 0.4 $\pm$ 0.0 & 67.6 \\
        DIKI & 69.4 $\pm$ 0.1 & 2.4 $\pm$ 0.0 & \textbf{42.5} \\
        LoRA-CL & 85.8 $\pm$ 0.0 & 0.4 $\pm$ 0.0 & 68.2 \\
        Tuna & 71.8 $\pm$ 0.2 & 10.4 $\pm$ 0.8 & 337.0 \\
        T2T-Finetune & 86.6 $\pm$ 2.3 & - & - \\
        \midrule
        \cellcolor{highlight}Our CHEEM & \cellcolor{highlight}84.5 $\pm$ 1.3 & \cellcolor{highlight}0.4 $\pm$ 0.0 & \cellcolor{highlight}63.0 \\
        \bottomrule
    \end{tabular}}
    \label{tab:clip-vdd}\vspace{-2mm}
\end{table}

\section{Smaller model for Task ID Recognition}
\label{sec:task-id-small-model}
Tables \ref{tab:mtil-small-taskid} and \ref{tab:vdd-small-taskid} present the results of using a smaller frozen backbone for task ID recognition alongside a larger backbone for final class prediction. For a fair comparison, we also include MoEAdapter4CL \cite{moeadapters4cl} as a baseline. MoEAdapters4CL uses a pretrained AlexNet backbone for task identification by training an autoencoder per task in the feature space, and infers the task ID at test time using the reconstruction error.

The results show that CHEEM can effectively decouple task identification and classification by using a lighter model (AlexNet or DEiT-Tiny) for task ID prediction and a stronger model (ViT-Base) for final classification. While using AlexNet for task identification leads to a noticeable drop in accuracy, replacing it with DEiT-Tiny results in only a negligible performance degradation. In fact, combining DEiT-Tiny for task prediction with ViT-Base for classification achieves performance comparable to using ViT-Base for both tasks, while significantly reducing computational cost (FLOPs).

Furthermore, CHEEM consistently outperforms MoEAdapters4CL when using either AlexNet or DEiT-Tiny for task identification. This demonstrates that CHEEM successfully leverages lightweight models for efficient task recognition without sacrificing overall performance, effectively combining efficiency and accuracy.

\begin{table}[h]
\centering
\caption{Comparison of Average Accuracy and Forgetting
\textbf{on the MTIL benchmark} with three seeds.}
\label{tab:mtil-small-taskid}

\resizebox{0.6\linewidth}{!}{
\begin{tabular}{r|c|c|c|c}
\toprule
\textbf{Method} & \textbf{Task ID} & \textbf{Avg. Acc} & \textbf{Avg. Frgt.} & \textbf{FLOPs} \\
\toprule

\multicolumn{5}{c}{\cellcolor{vitb}\textbf{ViT-Base}} \\
\midrule
MoEAdapter4CL & AlexNet & 71.5 $\pm$ 3.3 & 7.4 $\pm$ 1.2 & 35.4 \\
\cellcolor{highlight}CHEEM & ViT-Base & \cellcolor{highlight}\textbf{85.9} $\pm$ 0.3 & \cellcolor{highlight}1.7 $\pm$ 0.1 & \cellcolor{highlight}62.3 \\
\cellcolor{highlight}CHEEM$_{\text{lite}}$ & AlexNet & \cellcolor{highlight}79.9 $\pm$ 0.2 & \cellcolor{highlight}4.6 $\pm$ 0.2 & \cellcolor{highlight}34.1 \\
\cellcolor{highlight}CHEEM$_{\text{lite}}$ & DEiT-Tiny & \cellcolor{highlight}84.2 $\pm$ 0.5 & \cellcolor{highlight}2.1 $\pm$ 0.1 & \cellcolor{highlight}35.0 \\

\midrule
\multicolumn{5}{c}{\cellcolor{deit}\textbf{DEiT-Tiny}} \\
\midrule
MoEAdapter4CL & AlexNet & 53.7 $\pm$ 1.6 & 9.1 $\pm$ 1.3 & 3.7 \\
\cellcolor{highlight}CHEEM & DEiT-Tiny & \cellcolor{highlight}\textbf{74.5} $\pm$ 0.3 & \cellcolor{highlight}1.9 $\pm$ 0.0 & \cellcolor{highlight}4.5 \\
\cellcolor{highlight}CHEEM$_{\text{lite}}$ & AlexNet & \cellcolor{highlight}70.8 $\pm$ 0.6 & \cellcolor{highlight}4.1 $\pm$ 0.2 & \cellcolor{highlight}3.5 \\

\bottomrule
\end{tabular}}
\end{table}

\begin{table}[h]
\centering
\caption{Comparison of Average Accuracy and Forgetting
\textbf{on the VDD benchmark} with three seeds.}
\label{tab:vdd-small-taskid}

\resizebox{0.6\linewidth}{!}{
\begin{tabular}{r|c|c|c|c}
\toprule
\textbf{Method} & \textbf{Task ID} & \textbf{Avg. Acc} & \textbf{Avg. Frgt.} & \textbf{FLOPs} \\
\toprule

\multicolumn{5}{c}{\cellcolor{vitb}\textbf{ViT-Base}} \\
\midrule
MoEAdapter4CL & AlexNet & 70.2 $\pm$ 4.6 & 0.5 $\pm$ 0.1 & 35.4 \\
\cellcolor{highlight}CHEEM & ViT-Base & \cellcolor{highlight}\textbf{86.7} $\pm$ 0.2 & \cellcolor{highlight}0.4 $\pm$ 0.0 & \cellcolor{highlight}61.6 \\
\cellcolor{highlight}CHEEM$_{\text{lite}}$ & AlexNet & \cellcolor{highlight}83.8 $\pm$ 0.1 & \cellcolor{highlight}2.0 $\pm$ 0.1 & \cellcolor{highlight}28.8 \\
\cellcolor{highlight}CHEEM$_{\text{lite}}$ & DEiT-Tiny & \cellcolor{highlight}85.6 $\pm$ 0.3 & \cellcolor{highlight}1.1 $\pm$ 0.0 & \cellcolor{highlight}29.3 \\

\midrule
\multicolumn{5}{c}{\cellcolor{deit}\textbf{DEiT-Tiny}} \\
\midrule
MoEAdapter4CL & AlexNet & 65.6 $\pm$ 0.5 & 1.6 $\pm$ 0.1 & 3.7 \\
\cellcolor{highlight}CHEEM & DEiT-Tiny & \cellcolor{highlight}\textbf{76.18} $\pm$ 0.1 & \cellcolor{highlight}1.0 $\pm$ 0.0 & \cellcolor{highlight}4.5 \\
\cellcolor{highlight}CHEEM$_{\text{lite}}$ & AlexNet & \cellcolor{highlight}76.9 $\pm$ 0.2 & \cellcolor{highlight}1.0 $\pm$ 0.1 & \cellcolor{highlight}3.1 \\

\bottomrule
\end{tabular}}
\end{table}

\twocolumn

\section{Experiment Details}
\label{sec:training-and-data}

\textbf{Pretrained Models}: We initialize the pretrained ViT-B/16 and DEiT-Tiny/16 models from the checkpoint available in {\tt timm}. Both models use a patch size of 16 and a resolution of $224 \times 224$. The ViT-B/16 checkpoint has been pretrained on ImageNet 21k and finetuned on ImageNet1k. The DEiT-Tiny/16 checkpoint has been trained on ImageNet1k. All our experiments use the same checkpoints. We refer readers to \cite{vit} for the architecture details of ViT-B/16 and \cite{deit} for the architecture details of DEiT-Tiny/16.

Our experiments are conducted using PyTorch and leverage {\tt timm} for architecture implementation. In all our experiments, we use the Adam optimizer \cite{adam} with no weight decay. For experiments with CHEEM, we use a learning rate of $0.001$, 50 epochs for the supernet training and 20 epochs for finetuning. During supernet training, we use an exploration probability of $\epsilon = 0.3$, and use $\epsilon = 0.5$ during the target network selection to encourage more exploration. We do not perform any data augmentations, and simply resize the images to $224 \times 224$. We adapt the implementation from \url{https://github.com/GT-RIPL/CODA-Prompt} to perform experiments on CODA-Prompt, DualPrompts and L2P, and use our own implementations for the other baseline methods. We use a single Nvidia A100 GPU for all our experiments.

\subsection{Details of the MTIL benchmark}
The MTIL benchmark \cite{mtil} consists of 11 tasks: FGVC-Aircraft \cite{aircraft}, Caltech101 \cite{caltech-101}, CIFAR100 \cite{cifar}, Describable Textures \cite{dtd}, EuroSAT \cite{eurosat}, VGG-Flowers \cite{vgg-flowers}, Food101 \cite{food-101}, MNIST \cite{mnist}, Oxford Pets \cite{pets}, Stanford Cars \cite{stanfordcars}, SUN397 \cite{sun}. We use the official training and testing splits provided in the constituent datasets. We use the official validation splits for the evolutionary search, and create our own splits when official split is not provided by randomly sampling 10\% of the training dataset.

\begin{table}[H]
    \centering
    \caption{Number of samples in the training, validation, and test sets used in the experiments on the MTIL benchmark, along with the number of categories.}
    \resizebox{0.47\textwidth}{!}{
    \begin{tabular}{c|c|c|c|c}
    \toprule
         Task & \#Train & \#Validation & \#Test & \#Classes \\
         \midrule
         FGVC Aircraft & 3334 & 3333 & 3333 & 100 \\
         Caltech101 & 5465 & 608 & 2604 & 101 \\
         CIFAR100 & 45000 & 5000 & 19850 & 100 \\
         Describable Textures & 1880 & 1880 & 1880 & 47 \\
         EuroSAT & 17010 & 1890 & 8100 & 10 \\
         VGG-Flowers & 1020 & 1020 & 6149 & 102 \\
         Food-101 & 68175 & 7575 & 25250 & 101 \\
         MNIST & 54000 & 6000 & 10000 & 10 \\
         Oxford Pets & 3312 & 368 & 3669 & 37 \\
         Stanford Cars & 7329 & 815 & 8041 & 196 \\
         SUN397 & 17865 & 1985 & 19850 & 397 \\
         \bottomrule
    \end{tabular}}
    \label{tab:my_label}
\end{table}

\subsection{Details of the VDD benchmark}
The VDD benchmark \cite{vdd} consists of 10 tasks: ImageNet-1k~\citep{imagenet}, CIFAR100~\citep{cifar}, SVHN~\citep{svhn}, UCF101 Dynamic Images (UCF)~\citep{ucf1,ucf2}, Omniglot~\citep{omniglot}, German Traffic Signs (GTSR)~\citep{gtsrb}, Daimler Pedestrian Classification (DPed)~\citep{daimlerpedcls},  VGG Flowers~\citep{vgg-flowers}, FGVC-Aircraft~\citep{aircraft},    and Describable Textures (DTD)~\citep{dtd}. All the images in the VDD benchmark have been scaled such that the shorter side is 72 pixels. However, for a more realistic evaluation, we reconstruct the VDD benchmark with the original images and splits. Except for UCF101, Omniglot, and Daimler Pedestrian Classification, we use the official train, validation and test splits (when a validation split is not avaiable, we construct a validation split by randomly sampling 10\% of the training data.). Due to a lack of high resolution images for UFC101, Omniglot, and Daimler Pedestrian Classification, we use the splits and the images provided by the VDD benchmark and resize the images to $224 \times 224$.

\begin{table}[H]
    \centering
    \caption{Number of samples in the training, validation, and test sets used in the the experiments on the VDD benchmark, along with the number of categories.}
    \resizebox{\linewidth}{!}{
    \begin{tabular}{c|c|c|c|c}
  \toprule
  Task & \#Train & \#Validation & \#Test & \#Classes \\
  \midrule
  ImageNet12 & 1108951 & 123216 & 49000 & 1000 \\
  CIFAR100 & 45000 & 5000 & 19850 & 100 \\
  SVHN & 65931 & 7326 & 26032 & 10\\
  UCF & 6827 & 758 & 1952 & 101 \\
  Omniglot & 16068 & 1785 & 6492 & 1623 \\
  GTSR & 23976 & 2664 & 12630 & 43 \\
  DPed & 21168 & 2352 & 5880 & 2 \\
  VGG-Flowers & 1020 & 1020 & 6149 & 102 \\
  FGVC Aircraft & 3334 & 3333 & 3333 & 100 \\
  Describable Textures & 1880 & 1880 & 1880 & 47 \\
  \bottomrule
\end{tabular}}
\end{table}

\section{Theoretical Analysis of Local vs.\ Global Argmax of Head Classifiers in Continual Learning}\label{sec:growing-head-analyses}
As seen in Section \ref{sec:performance}, CODA-Prompt, DualPrompt and L2P perform significantly worse than LoRA-C and CHEEM. This large drop is attributed to the discrepancy between local and global softmax. We verify this in Table \ref{tab:global-vs-local-performance}, which shows that when provided with a task ID to retrieve the appropriate local part of the head during inference, the performance of CODA-Prompt, DualPrompt and L2P is significantly better, almost approaching S-Prompts and DIKI. We provide a theoretical analysis in the following sections.

\begin{table*}[t]
    \centering
    \caption{Acc$_{Global}$ refers to the average accuracy (Eqn. \ref{eq:avg_accuracy_lll}) calculated using the global head, and Acc$_{Local}$ refers to the same but by masking the logits not belonging to the task. Acc$_{Train}$ refers to the accuracy calculated after the training on a task is complete, averaged over all the tasks.}
    \label{tab:global-vs-local-performance}
    \resizebox{0.9\textwidth}{!}{
    \begin{tabular}{c|c|c|c|c|c|c}
    \toprule
        & \multicolumn{3}{c|}{\cellcolor{vitb}\textbf{ViT-B}} & \multicolumn{3}{c}{\cellcolor{deit}\textbf{DEiT-Tiny}} \\
        \midrule
        \textbf{Method} & \textbf{Acc}$_{Global}$ & \textbf{Acc}$_{Local}$ & \textbf{Acc}$_{Train}$ & \textbf{Acc}$_{Global}$ & \textbf{Acc}$_{Local}$ & \textbf{Acc}$_{Train}$ \\
        \midrule
        CODA-Prompt & 40.22 $\pm$ 1.22 & 79.70 $\pm$ 0.61 & 86.18 $\pm$ 0.02 & 5.62 $\pm$ 0.25 & 34.72 $\pm$ 1.62 & 67.53 $\pm$ 0.37 \\
        DualPrompt & 33.82 $\pm$ 0.35 & 83.61 $\pm$ 0.13 & 84.63 $\pm$ 0.09 & 30.89 $\pm$ 0.29 & 68.17 $\pm$ 0.24 & 71.25 $\pm$ 0.10  \\
        L2P & 26.61 $\pm$ 0.16 & 80.03 $\pm$ 0.58 & 84.95 $\pm$ 0.11 & 23.24 $\pm$ 0.14 & 60.79 $\pm$ 0.67 & 71.47 $\pm$ 0.08 \\
        S-Prompts & 81.62 $\pm$ 0.35 & 84.48 $\pm$ 0.18 & 84.48 $\pm$ 0.18 & 67.33 $\pm$ 0.38 & 70.71 $\pm$ 0.40 & 70.71 $\pm$ 0.40 \\
        DIKI & 76.42 $\pm$ 0.04 & 84.50 $\pm$ 0.04 & 84.50 $\pm$ 0.04 & 67.63 $\pm$ 0.06 & 70.86 $\pm$ 0.07 & 70.86 $\pm$ 0.07 \\
        CHEEM & 85.88 $\pm$ 0.29 & 88.68 $\pm$ 0.16 & 88.68 $\pm$ 0.16 & 74.51 $\pm$ 0.28 & 78.11 $\pm$ 0.31 & 78.11 $\pm$ 0.31 \\
        \bottomrule
    \end{tabular}}
    \label{tab:acc-head-analysis}
\end{table*}

\subsection{The problem}
In continual learning, we have $N$ tasks, each with a different number of classes. Let task $t$ have $C_t$ classes, so by time $T$ we have observed tasks $1,\dots,T$ with a total of $\sum_{t=1}^T C_t$ classes. We train a shared feature extractor $\phi(\mathbf{x}) \in \mathbb{R}^d$ and a growing head classifier composed of task-specific segments $W^t \in \mathbb{R}^{d \times C_t}$. 

During training of task $t$, only the segment $W^t$ is updated and used in a softmax over the $C_t$ classes for the current task. However, at inference, for a new test sample $\mathbf{x}$ belonging (in truth) to task $t^*$, the entire head is used: we compute logits for \emph{all} classes seen so far, and choose the global $\arg\max$. We denote:

\begin{itemize}
    \item \emph{Local argmax}:
    \[
      \hat{y}_{\text{local}}(\mathbf{x})
      \;=\;
      \arg\max_{c \in \{1,\dots,C_{t^*}\}} \; z_{t^*,c}(\mathbf{x}),
    \]
    where $z_{t^*,c}(\mathbf{x}) = \langle W^t_{(\cdot,c)}, \phi(\mathbf{x}) \rangle$ are the logits restricted to task $t^*$.
    \item \emph{Global argmax}:
    \[
      \hat{y}_{\text{global}}(\mathbf{x})
      \;=\;
      \arg\max_{(t,c) \in \{1,\dots,T\} \times \{1,\dots,C_t\}} \; z_{t,c}(\mathbf{x}).
    \]
\end{itemize}

We are interested in the probability that these two predictions coincide:
\[
  \Pr\bigl(\hat{y}_{\text{local}}(\mathbf{x}) \;=\; \hat{y}_{\text{global}}(\mathbf{x})\bigr).
\]
Below is a stylized theoretical analysis of why and how often these two can match, highlighting the factors that influence this probability.

\subsection{Distribution of Logits and Task Separation}
Let $z_{t,c}(\mathbf{x})$ be the logit for class $c$ in task $t$ for sample $\mathbf{x}$. We may approximate $z_{t,c}(\mathbf{x})$ by a random variable with mean $\mu_{t,c}$ and variance $\sigma_{t,c}^2$, e.g.,
\[
  z_{t,c}(\mathbf{x}) \;\approx\; \mu_{t,c} + \epsilon_{t,c}, \quad 
  \epsilon_{t,c} \sim \mathcal{N}(0, \sigma_{t,c}^2).
\]
In reality, these means and variances depend on how well the feature $\phi(\mathbf{x})$ and the weights $W^t$ are aligned, but we treat them as parameters to illustrate.

Define:
\begin{align}
  & \max_{c \in C_{t^*}} z_{t^*,c}(\mathbf{x}) 
  \;\;\text{(the local max for the correct task)},
  \\
  & \max_{(t\neq t^*)} \max_{c \in C_t} \; z_{t,c}(\mathbf{x})
  \;\;\text{(the max out-of-task logit)}.
\end{align}
For $\hat{y}_{\text{local}} = \hat{y}_{\text{global}}$, we need 
\[
  \max_{c \in C_{t^*}} z_{t^*,c}(\mathbf{x}) 
  \;\;\ge\;\; 
  \max_{(t \neq t^*)}\;\max_{c} \; z_{t,c}(\mathbf{x}).
\]
Hence the distribution of all out-of-task logits relative to the best in-task logit is crucial.

\subsection{Probability of Matching Local and Global Argmax}
\subsubsection{A Basic Two-Class Example}
Consider just one class $c^*$ in the true task vs.\ one class $k$ in an \emph{other} task. Suppose
\[
  z_{t^*, c^*} \sim \mathcal{N}(\mu^*, \sigma^2),
  \quad
  z_{t', k} \sim \mathcal{N}(\mu', \sigma^2).
\]
The probability that $z_{t^*,c^*} \ge z_{t',k}$ is
\[
  \Pr(z_{t^*,c^*} \ge z_{t',k})
  = \Pr(z_{t^*,c^*} - z_{t',k} \ge 0)
  = \Phi\!\Bigl(\frac{\mu^* - \mu'}{\sqrt{2}\,\sigma}\Bigr),
\]
where $\Phi$ is the standard normal CDF.

\subsubsection{Many Classes from Different Tasks}
Now suppose there are $C_{t^*}$ classes in the correct task, and $M = \sum_{t \neq t^*} C_t$ classes outside. Let the local maximum
\[
  Z^* \;=\; \max_{c \in \{1,\dots,C_{t^*}\}} \; z_{t^*,c},
\]
and let $Z_1,\dots,Z_M$ represent the logits of the $M$ out-of-task classes. Then
\[
  \Pr(\hat{y}_{\text{local}} = \hat{y}_{\text{global}})
  \;=\;
  \Pr\Bigl(Z^* \;\ge\; \max\{Z_1,\dots,Z_M\}\Bigr).
\]
If $Z^*$ is (roughly) $\mathcal{N}(\mu_{\text{local}}, \sigma_{\text{local}}^2)$ and each $Z_j$ is $\mathcal{N}(\mu_o, \sigma_o^2)$ (independent simplification), then
\[
  \Pr\bigl(Z^* \ge Z_j \text{ for all } j\bigr)
  \;=\;
  \int 
    \Bigl[\Pr(Z_j \le z)\Bigr]^M
    \, F_{Z^*}(z) \, dz.
\]
When $\mu_{\text{local}} > \mu_o$, this probability is high for moderate $M$, but as $M$ grows, the chance that \emph{some} out-of-task class logit exceeds $Z^*$ increases, unless the gap $\mu_{\text{local}} - \mu_o$ is large.

\subsection{Factors Influencing the Match Probability}
\begin{enumerate}
    \item \textbf{Feature Separation Across Tasks.} If $\phi(\mathbf{x})$ strongly separates tasks, then for $\mathbf{x}$ from task $t^*$, out-of-task logits $z_{t,c}$ for $t \neq t^*$ are consistently lower. This increases the probability of $\hat{y}_{\text{local}} = \hat{y}_{\text{global}}$.
    \item \textbf{Logit Magnitude \& Variance.} Even if the \emph{means} of the correct task's logits exceed those of other tasks, high variance or overlap can cause out-of-task classes to occasionally exceed the correct task's maximum.
    \item \textbf{Regularization and Task Order.} Continual-learning methods that regularize old task weights or use replay data reduce the chance of weight drift, making it less likely that earlier or other tasks overshadow the correct one.
    \item \textbf{Task Size Differences.} Larger tasks (more classes) or tasks that were trained earlier might have stronger classifier weights. Conversely, smaller tasks might have very tight, well-separated features. Both can affect how likely a mismatch is.
\end{enumerate}

\subsection{A Rough Illustrative Bound}
As a simplistic illustration, suppose:
\begin{itemize}
    \item For task $t^*$, the local maximum logit $Z^*$ has mean $\mu^*$ and variance $\sigma^{*2}$.
    \item All out-of-task classes have means $\mu_o < \mu^*$ and variance $\sigma_o^2$.
    \item There are $M$ out-of-task classes in total.
\end{itemize}
Then
\[
  \Pr(\hat{y}_{\text{local}} = \hat{y}_{\text{global}})
  \;\approx\;
  \int
    \Bigl[\Pr(Z_o \le z)\Bigr]^{M} 
    F_{Z^*}(z) 
  \,dz,
\]
where $Z_o$ is the logit distribution for a single out-of-task class and $F_{Z^*}$ is the PDF of $Z^*$. If $\mu^*$ is sufficiently larger than $\mu_o$ (and variances are not too large), $Z^*$ will, with high probability, exceed \emph{all} $M$ out-of-task logits. But as $M$ grows large, this event can become less likely unless the margin $\mu^* - \mu_o$ is also large.

\subsection{Remarks}
Overall, the probability that the local argmax (over the correct task only) coincides with the global argmax (over all tasks/classes) depends on:
\begin{itemize}
    \item How well the feature extractor $\phi$ separates tasks, so that out-of-task logits stay low for samples of task $t^*$.
    \item The relative scale and calibration of classifier weights $W^t$ across tasks.
    \item The total number of classes from other tasks that could ``compete'' and produce a large logit by chance.
\end{itemize}
If tasks are well-separated (and the classifier is carefully regularized or calibrated), this probability can be very high. Conversely, if many classes from older or different tasks produce comparably large logits, the global $\arg\max$ may differ from the local $\arg\max$ more frequently as the number of tasks and classes increases.

\section{Identifying the Task-Synergy Internal Memory in ViTs}
\label{sec:place_cheem}

The left of Fig.~\ref{fig:flow} shows a ViT block. Denote by $x_{L,d}$ an input sequence consisting of $L$ tokens encoded in a $d$-dimensional space. In ViTs, the first token is the so-called class-token, {\tt CLS}. The remaining $L-1$ tokens are formed by patchifying an input image and then embedding patches, together with additive positional encoding. A ViT block is defined by, 
\begin{align}
    z_{L, d} & = x_{L, d} + \text{Proj}\Bigl(\text{MHSA}\bigl(\text{LN}_1(x_{L, d})\bigr)\Bigr), \label{eq:mhsa_proj} \\
    y_{L, d} & = z_{L, d} + \text{FFN}\Bigl(\text{LN}_2(z_{L, d})\Bigr), \label{eq:ffn}
\end{align}
where $\text{LN}(\cdot)$ represents the layer normalization~\citep{ba2016layer}, and $\text{Proj}(\cdot)$ is a linear transformation fusing the multi-head outputs from MHSA module. 
The MHSA realizes the dot-product self-attention between Query and Key, followed by aggregating with Value, where Query/Key/Value are linear transformatons of the input token sequence. The FFN is often implemented by a multi-layer perceptron (MLP) with a feature expansion layer $\text{MLP}^{\text{Up}}$ and a feature reduction layer $\text{MLP}^{\text{Down}}$ with a nonlinear activation function (such as the GELU~\citep{hendrycks2016gaussian}) in the between, i.e., $\text{FFN}(\cdot)=\text{MLP}^{\text{Down}}\Bigl(\text{GELU}\bigl(\text{MLP}^{\text{Up}}(\cdot)\bigr)\Bigr)$.

\begin{table}[t]
    \centering
    \caption{Ablation studies of identifying where to place our proposed CHEEM in ViT by testing 11 components or composite components (Eqns.~\ref{eq:mhsa_proj} and~\ref{eq:ffn}). 
    }
    \resizebox{0.45\textwidth}{!}{
        \begin{tabular}{l|l|c|c}
            \toprule
            \textbf{Index} & \textbf{Finetuned Component} &  \textbf{Avg. Acc.} & \textbf{Avg. Forgetting} \\
            \midrule
            1 & $\text{LN}_1$ + $\text{LN}_2$ & $81.76$ & $21.24$ \\
            \midrule
            2 & $\text{FFN}$ & $84.20$ & $44.76$ \\
            \rowcolor{highlight} 3 & $\text{MLP}^{\text{Down}}$ & $83.66$ & $37.99$ \\
            4 & $\text{LN}_2$ & $80.04$ & $16.35$ \\
            \midrule
            5 & MHSA + $\text{LN}_1$ & $85.26$ & $54.38$ \\
            6 & LN$_1$ & $81.18$ & $19.04$ \\
            7 & Query & $81.57$ & $19.69$ \\
            8 & Key & $81.56$ & $19.19$ \\
            9 & Query+Key & $81.49$ & $31.10$ \\
            10 & Value & $84.99$ & $37.58$ \\
            \rowcolor{highlight}  11 & Projection  & $85.11$ & $30.50$ 
            \\
            \bottomrule
        \end{tabular}
    }
    \label{tab:acc_vs_forgetting} \vspace{-3mm}
\end{table}

The proposed identification process is straightforward. Without introducing any modules handling forgetting, we compare both the task-to-task forward transferrability and  the sequential forgetting for different components in a ViT block. \textbf{Our intuition is that a desirable component for placing the task-synergy parameter memory must enable strong transferrability with manageable forgetting, while being lightweight to account for the trade-off between stability and plasticity.}

To that end, we use the VDD benchmark~\citep{vdd} (see Fig.~\ref{fig:teaser-vdd}). We first train a ViT-Base~\citep{vit} on the first task, ImageNet~\citep{imagenet}, as the base model $F_1(\cdot)$. 
To measure the task-to-task transferability, we \textit{individually fine-tune}  $F_{1}$ in a task-to-task transfer learning  manner for the remaining 9 streaming tasks.
Let $F_{t|1}$ be the backbone fine-tuned for  task $T_t$ (for $t\geq 1$), and $C_t$ the  head classifier trained from scratch. The average Top-1 accuracy is defined by Equation \ref{eq:avg_accuracy_lll}
where $\text{Acc}()$ uses the Top-1 classification accuracy. %

To measure the sequential forgetting, we \textit{continually fine-tune} the backbone started from $F_1$ on the 9 tasks in a randomly sampled and fixed streaming order (as shown in Fig.~\ref{fig:teaser-mtil} in the main text). Let $F_{1:t}$ be the backbone trained sequentially and continually after task $T_t$ and $H_t$ is its head classifier. 
The average forgetting~\citep{riemannian-walk} on the first $N-1$ streaming tasks is defined by Equation \ref{eq:avg_forgetting},
where $a_{j,t}=\text{Acc}(T_t; F_{1:j}, H_t)$.

As shown in Table~\ref{tab:acc_vs_forgetting}, we compare 11 components or composite components in ViT. 
Consider the strong forward transfer ability, manageable forgetting, maintaining simplicity and for less invasive implementation in practice, \textbf{we select either the Projection layer after the MHSA or the MLP$^\text{Down}$ as the task-synergy internal (parameter) memory}  to realize our proposed CHEEM for ExfCCL (Fig.~\ref{fig:flow}). We test both in experiments and provide ablation studies in Section \ref{sec:ablation-studies}.